\documentclass[sigconf,natbib=true,anonymous=false]{acmart}
\usepackage{graphicx}
 
\usepackage{multirow}
\usepackage[a-1b]{pdfx} 
\usepackage[ruled,vlined]{algorithm2e}
\usepackage{balance}
\usepackage{xcolor}
\usepackage{colortbl}
\usepackage{subfigure}
\usepackage{comment}
\definecolor{Gray}{gray}{0.65}
\AtBeginDocument{%
  \providecommand\BibTeX{{%
    \normalfont B\kern-0.5em{\scshape i\kern-0.25em b}\kern-0.8em\TeX}}}

\setcopyright{acmcopyright}
\copyrightyear{2018}
\acmYear{2018}
\acmDOI{XXXXXXX.XXXXXXX}

\acmConference[MLG '23]{19th International Workshop on Mining and Learning with Graphs}{Aug 09,
 2023}{Long Beach, CA, USA}
\acmPrice{15.00}
\acmISBN{978-1-4503-XXXX-X/18/06}

\begin{document}

\title{Seq-HyGAN: Sequence Classification via Hypergraph Attention Network}

\author{Khaled Mohammed Saifuddin}
\orcid{0000-0002-0903-937X}
\authornotemark[1]
\affiliation{%
  \institution{Georgia State University}
  \streetaddress{Downtown}
  \city{Atlanta}
  \state{Georgia}
  \country{USA}
  \postcode{30302}
}
\email{ksaifuddin1@student.gsu.edu}
\author{Corey May}
\affiliation{%
  \institution{Arkansas Tech University }
  \streetaddress{Downtown}
  \city{Russellville}
  \state{Arkansas}
  \country{USA}
  \postcode{72801}
}
\email{cmay12@atu.edu}

\author{Farhan Tanvir}
\affiliation{%
  \institution{Oklahoma State University}
  \streetaddress{Downtown}
  \city{Stillwater}
  \state{Oklahoma}
  \country{USA}
  \postcode{74078}
}
\email{farhan.tanvir@okstate.edu}

\author{Muhammad Ifte Khairul Islam}
\affiliation{%
  \institution{Georgia State University}
  \streetaddress{Downtown}
  \city{Atlanta}
  \state{Georgia}
  \country{USA}
  \postcode{30302}
}
\email{mislam29@student.gsu.edu}

\author{Esra Akbas}
\affiliation{%
  \institution{Georgia State University}
  \streetaddress{Downtown}
  \city{Atlanta}
  \state{Georgia}
  \country{USA}
  \postcode{30302}
}
\email{eakbas1@gsu.edu}

\renewcommand{\shortauthors}{Saifuddin et al.}

\newcommand{\sht}{\texttt{Seq-HyGAN}}
\begin{abstract}
  Extracting meaningful features from sequences and devising effective similarity measures are vital for sequence data mining tasks,
particularly sequence classification. While Neural Network models are commonly used to learn features of sequence automatically, they are limited to capturing adjacent structural connection information and ignore global, higher-order information between the sequences. To address these challenges, we propose a novel Hypergraph Attention Network model, namely \sht\, for sequence classification problems. To capture the complex structural similarity between sequence data, we create a novel hypergraph model by defining higher-order relations between subsequences extracted from sequences. 
 Subsequently, we introduce a Sequence Hypergraph Attention Network that learns sequence features by considering the significance of subsequences and sequences to one another. 
 Through extensive experiments, we demonstrate the effectiveness of our proposed \sht\ model in accurately classifying sequence data, outperforming several state-of-the-art methods by a significant margin. 
 This paper belongs to "Algorithms and methods - Graph neural networks and graph representation learning" and is in the "Novel research papers" category.
\end{abstract}




\maketitle

\section{Introduction}
Extracting meaningful features from sequences and devising effective similarity measures are vital for sequence data mining tasks, particularly sequence classification. 
Neural networks (NN), especially recurrent neural networks (RNN) (i.e., LSTM, GRU), are commonly used to learn features capturing the adjacent structural information. However, these models may struggle to capture non-adjacent, high-order, and complex relationships present in the sequence data. 
Recently, graph has also been explored for different types of sequence data classification tasks such as text classification~\cite{Saifuddin2021}, DNA-protein binding prediction \cite{Guo}, protein function prediction \cite{Gligorijevic2021}, drug-drug interaction prediction~\cite{Bumgardner2022}, etc. Graphs, as highly sophisticated data structures, have the capability to effectively capture both local and global non-adjacent information within the data \cite{Qian2019, islam2021proximity}. The state-of-the-art models for sequence-to-graph conversion can be classified into two main categories: order-based graphs and similarity-based graphs  \cite{compeau2011bruijn, atkinson2009using, zola2014constructing}. 

While existing graph models for sequence data achieved great performance, they still have some key challenges. Order-based models generate many large and sparse graphs, especially when dealing with a long and large number of sequences. Handling such many large graphs can lead to increased computational and memory requirements. Similarly, similarity-based graphs encounter challenges when calculating similarities between all pairs of sequences, especially for large datasets. Furthermore, order-based graphs fail to capture relationships between sequences beyond the intra-sequence connections with considering dyadic relationships between nodes. However, sequence data may possess more complex relationships, such as triadic or tetradic relationships, which these models are unable to capture. The selection of appropriate similarity measures becomes problematic, and using similarity values to define relationships between sequences can lead to information loss.

In order to overcome the aforementioned challenges for sequence classification, we propose a novel Hypergraph Attention Network model, namely \sht. Our approach is built on the assumption that sequences sharing structural similarities tend to belong to the same classes, and sequences can be considered similar if they contain similar subsequences. To effectively capture the structural similarities between sequences, we represent them in a hypergraph framework, where the sequences are depicted as hyperedges that connect their respective subsequences as nodes. This construction allows us to create a single hypergraph encompassing all the sequences in the dataset. 
Unlike a standard graph where the degree of each edge is 2, hyperedge is degree-free; it can connect an arbitrary number of nodes \cite{Bretto, Aktas2021, Aktas2021a}. To enhance the representation of sequences and capture complex relationships among them, we introduce a novel \sht\ architecture that employs a three-level attention-based neural network. Unlike regular NNs (i.e., RNN) that can only capture local information, \sht\ is more robust as it can capture both local (within the sequence) and global (between sequences) information from the sequence data. Moreover, while traditional Graph Neural Networks (GNNs) with standard graphs are limited to a message-passing mechanism between two nodes, this hypergraph setting assists GNNs in learning a much more robust representation of sequences with a message-passing mechanism not only between two nodes but between many nodes and also between nodes and hyperedges.

Our contributions are summarized as follows:
\begin{itemize}
\item \textbf{Hypergraph construction from sequences: } 
We introduce a novel hypergraph construction model from the sequence dataset. In our proposed sequence hypergraph, each subsequence extracted from the sequences is represented as a node, while each sequence, composed of a unique set of subsequences, is represented as a hyperedge. By leveraging this hypergraph model, we can capture and define higher-order complex structural similarities that exist between sequences.

\item  \textbf{Hypergraph Attention Network:}
{We introduce a novel hypergraph attention network model, referred to as \sht, specifically designed for sequence classification tasks. Our model focuses on learning sequence representations as hyperedges while considering both local and global context information with three levels of aggregation with attention that capture different levels of context. At the first level, it generates node embedding that incorporates global context by aggregating hyperedge embeddings. At the second level, the model refines node embeddings for each hyperedge. It captures local context by aggregating neighboring node embeddings in the same hyperedge and also considering the position of subsequences in a sequence. Finally, at the third level, it generates sequence embedding by aggregating node embeddings from both global (level 1) and local (level 2) perspectives, resulting in a comprehensive representation of the sequences.}

\item \textbf{Capturing importance via Attention:}{ To capture the relative importance of individual subsequences (nodes) within each sequence (hyperedge), our model incorporates an attention mechanism. This mechanism allows the model to learn the varying significance of specific subsequences in contributing to the overall similarity between sequences. Additionally, the attention mechanism enables the model to discern the importance of both subsequences and sequences in relation to each subsequence. By doing so, our model comprehensively captures the inter-dependencies between different levels of granularity in the data, facilitating a nuanced understanding of the structure of the data.}

\item  \textbf{Extensive experiments:} We conduct extensive experiments to show the effectiveness of our model on four different datasets and five different classification problems. We also compare the proposed \sht\ model with the state-of-the-art baseline models. The results with different accuracy measures show that our method significantly surpasses the baseline models.
\end{itemize}

The rest of this paper is organized as follows. In Section \ref{sec:works}, we provide a concise overview of the existing literature on sequence classification and hypergraph GNNs. Section \ref{sec:method} presents the proposed \sht\ model, outlining the process of constructing a hypergraph from a sequence dataset and the details of the proposed hypergraph attention network. Subsequently, Section \ref{sec:experiment} presents the results obtained from our extensive experiments. Finally, Section \ref{sec:con} concludes the paper by summarizing the key findings.

\vspace{-2mm}
\section{Related Work} \label{sec:works}
Various studies have been carried out on the problem of sequence classification. We broadly categorize them into three different types of methods: Machine Learning(ML)-based, Deep Learning (DL)-based, and GNN-based. ML-based methods generate a feature vector using some kernel function such as k-spectrum kernel~\cite{leslie2001spectrum}, local alignment kernel~\cite{Saigo2004}, then they apply ML classifiers for sequence classification tasks.
 \begin{figure*}[h!]
  \centering
  \subfigure[Hypergraph Construction]{\includegraphics[width=8.5cm, height=4cm]{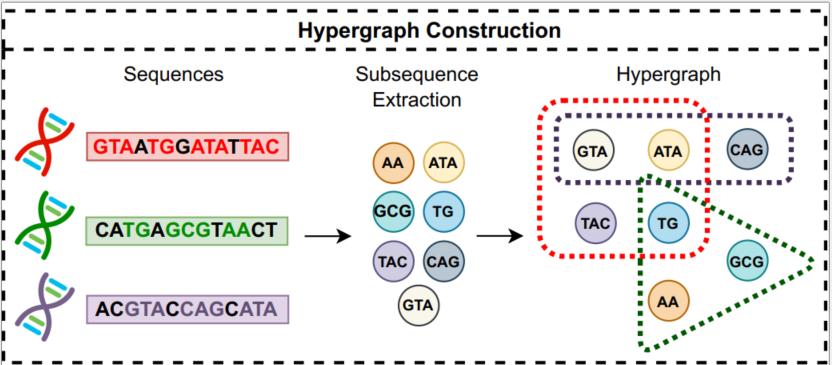}}
  \hfill
  \subfigure[\sht]{\includegraphics[width=8.5cm,height=4cm]{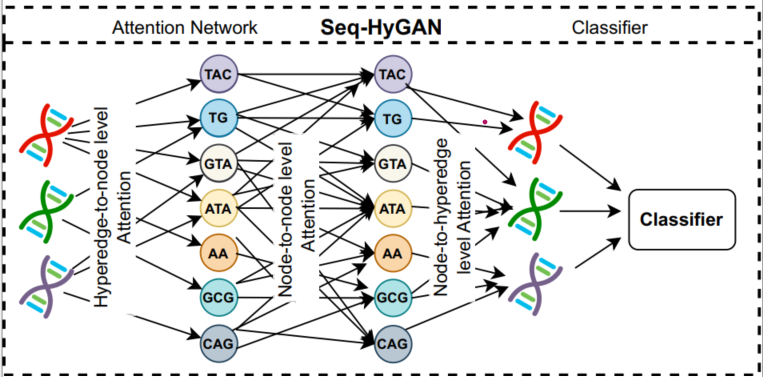}}\vspace{-3mm}
   \caption{System architecture of the proposed method. The first step is hypergraph construction, where each sequence (e.g., DNA) is a hyperedge, and the (frequent) subsequences of sequences are the nodes. The second step is an attention-based hypergraph neural network model for sequence classification, namely \sht, that generates the representations of sequences while giving more attention to the important subsequences.}
    \label{fig:method} \vspace{-2mm}
\end{figure*}  
Applications of DL-based methods are also pretty common in this domain~\cite{ Nguyen2016, Torrisi2020, Eickholt2013, Timmons2021}. While some studies use one type of DL method, some studies create hybrid models  by combining different DL methods. 
Network-based models have also been explored to analyze sequence data~\cite{Ashoor2020, Hwang2019}. A common approach for representing genome sequence in a network is to use the De-Bruijn graph~\cite{Turner}. To construct a De-Bruijn graph, the $k$-mer method is first applied to a sequence input. Each generated $k$-mer token is used as a node, and subsequent $k$-mers having an overlap in $k-1$ positions are connected using an edge to construct the graph. 

GNNs have exhibited great performances in different research fields, including graph convolutional network (GCN), which has been successfully applied for sequence data analysis~\cite{Huangc, Zhanga}. In \cite{Yaoa}, authors apply GCN for text classification by creating a heterogeneous text graph including document and word nodes from the whole corpus. The same architecture is applied for DNA-protein binding prediction from sequential data~\cite {Guo}. Their networks have sequence and token nodes extracted from sequences by applying $k$-mer.

Due to the ability to capture higher-order complex relations, hypergraph and hypergraph neural networks have recently gotten huge attention in different research domains \cite{Feng2018, kim, saifuddin2022hygnn}. These works mainly focus on node representation by using hypergraph neural networks. However, in our proposed \sht, we design a novel three-level attention network to get the representation of hyperedges instead of nodes. Moreover, to the best of our knowledge, this paper is the first to address the sequence classification problem using hypergraph and attention-based hypergraph neural networks.  
\vspace{-2mm}
\section{Methodology} \label{sec:method} 
\subsection{Preliminaries and settings}
\label{sec:PF}
Sequence classification is the problem of predicting the class of sequences. Our motivation in this work is that patterns as subsequences are important features of sequences, and if two sequences share many patterns, they have a higher similarity. Also, it is assumed that similar sequences will have the same class labels. To define the higher-order pattern-based similarity between sequences, we construct a hypergraph from the sequence data.  Below is a formal definition of a hypergraph.
\begin{definition}
{\textbf{Hypergraph:}}
A hypergraph is defined as ${G} = ({V}, {E})$ where ${V} =\{v_{1},...,v_{i}\}$ is the set of nodes and ${E} = \{e_{1},...,e_{j}\}$ is the set of hyperedges. Hyperedge is a special kind of edge that consists of any number of nodes. Similar to the adjacency matrix of a graph, a hypergraph can be denoted by an incidence matrix $H^{n \times m}$ where $n$ is the number of nodes, $m$ is the number of hyperedges and $H_{ij}=1$ if $v_i \in e_j$, otherwise 0.
\end{definition}
 After hypergraph creation, to accomplish the sequence classification task, we propose an attention-based hypergraph neural network model consisting of a novel three-level attention mechanism that learns the importance of the subsequences (nodes) and, thus, the representation of the sequences (hyperedges). We train the whole model in a semi-supervised fashion. Our proposed model has two steps:
(1) Hypergraph construction from the sequence data, (2) Sequence classification using attention-based hypergraph neural networks.

\subsection{Sequence Hypergraph Construction}\label{sec:HGC}
In order to capture the similarity between sequences, we define the relationship between sequences based on  common subsequences within each sequence. We represent this relationship as a hypergraph that captures the higher-order similarity of the sequences. First, we decompose the sequences into a set of subsequences as the important patterns of the sequences. Then, we represent this set of subsequences as the nodes of the hypergraph, and each sequence including a set of subsequences is a hyperedge. Each hyperedge may connect with other hyperedges through some shared nodes as subsequences. Thus, this constructed hypergraph defines a higher-level connection of sequences and subsequences and helps to capture the complex similarities between the sequences. Moreover, this hypergraph setting ensures a better robust representation of sequences with a GNN model having a message-passing mechanism not only limited to two nodes but rather between the arbitrary number of nodes and also between edges through nodes. The steps of hypergraph construction are shown in Algorithm~\ref{algorithm:1}.


Different algorithms can be used to generate the subsequences for hypergraph construction, such as ESPF \cite{huang2019explainable}, $k$-mer \cite{zhang2017mining}, strobemers \cite{sahlin2021strobemers}. In this paper, we exploit ESPF and $k$-mer to generate subsequences and examine their effects on the final results. 
While $k$-mer uses all the extracted subsequences for a certain $k$ value, ESPF only selects the most frequent subsequences from a list of candidate subsequences for a certain threshold. Appendix \ref{subsequence} provides comprehensive information on ESPF and $k$-mer, including their respective algorithms.

\subsection{ Sequence Hypergraph Attention Network}\label{sec:HYGNN}
Sequence classification is the problem of predicting the class of sequences. To classify sequences, it is essential to generate feature vectors that can effectively embed the structural information. Since in our hypergraph model, we represent each sequence as a hyperedge; we need to learn hyperedge representation. Regular GNN models that generate the embedding of nodes do not work on our hypergraph. Therefore, we propose a novel Sequence Hypergraph Attention Network model, namely \sht.
\SetKwComment{Comment}{/* }{ */} 
\begin{algorithm}[t!]
\SetAlgoLined
\textbf{Input: } $Sequences$\\
 \textbf{Output: }Hypergraph incident matrix: $H$\\

Subsequence\_list<- $Sequence\_Decomposition(Sequences)$\Comment*[r]{Sequence\_Decomposition() could be ESPF or $k$-mer that decomposes sequences into moderated size subsequences.}
 \For {each subsequence in Subsequence\_list}
 {
      \If {subsequence is in Sequence\_dictionary[sequence]}{
      $H[i,j]=1$ \Comment*[r]{i,j is the id of subsequence and sequence, respectively.} }
 }
\textbf{Output: } Hypergraph incident matrix, $H$
\caption{Sequence Hypergraph Construction }
\label{algorithm:1}
\end{algorithm} 
Given the $f$ dimensional hyperedge feature matrix, $X \in R^{E \times f}$ and incidence matrix $H \in R^{V \times E}$, \sht\ generates a hyperedge feature vector of $f'$ dimension via learning a function $F$. Then it predicts labels for sequences using generated feature vectors.   Our proposed model leverages memory-efficient self-attention mechanisms to capture high-order relationships in the data while preserving both local and global context information. It comprises a three-level attention network: hyperedge-to-node, node-to-node, and node-to-hyperedge levels. At the hyperedge-to-node level, attention is utilized to aggregate hyperedge information and generate node representations that encapsulate global context. The node-to-node level attention refines the node representations by aggregating information from neighboring nodes within the same hyperedge, capturing local context. Lastly, the node-to-hyperedge level attention aggregates node representations from both local and global context perspectives to generate hyperedge representations with attention.
We define tree attention layers in general as follows.
\\
\begin{align}
p_{i}^{l}=\mathbf{AG_{E-V}}^{l}(p_{i}^{l-1},{n_{j}^{l-1}}|\forall e_{j}\in E_{i}),\\
m_{i,j}^{l}=\mathbf{AG_{V-V}}^{l}(p_{i}^{l},p_{y}^{l}| \forall v_{y}\in e_{j}),\\
n_{j}^{l}=\mathbf{AG_{V-E}}^{l}(n_{j}^{l-1},{p_{i}^{l}},{m_{i,j}^{l}}|\forall v_{i}\in e_{j})
\end{align}
where \textbf{AG\textsubscript{E-V}} (Hyperedge-to-Node) aggregates the information $n_j$ of all hyperedges $e_j$ to generate the $l$-th layer representation $p_{i}^l$ of node $v_i$ and $E_{i}$ is the set of hyperedges that node $v_i$ belongs to. \textbf{AG\textsubscript{V-V}}(Node-to-Node) generate the $l$-th layer representation $m_{i,j}^l$ of node $v_i$ for a specific hyperedge $e_j$ by aggregating all the nodes $v_{y}$ present in $e_j$.  Finally, \textbf{AG\textsubscript{V-E}}(Node-to-Hyperedge) aggregates the information of all nodes $v_{i}$ that belongs to hyperedge $e_j$ to generate the $l$-th layer representation $n_{j}^l$ of $e_j$.
\\
{\textbf{Hyperedge-to-node level attention:}} As our first layer, we aggregate information from hyperedges to learn the representation of nodes to capture the global context in the hypergraph. Although a node may belong to different hyperedges, all hyperedges may not be equally important for that node. To learn the importance of hyperedges and incorporate it into the representation of nodes, we design a self-attention mechanism. While aggregating hyperedge representations for a node, this attention mechanism ensures more weight to important hyperedges than others. With the attention mechanism, the $l$-th layer node feature $p_{i}^l$ of node $v_i$ is defined as
 
\begin{equation}
\label{eqn_n_4}
p_{i}^{l}=\alpha \left(\sum_{e_{j}\in {E_{i}}}\Gamma_{ij}W_{1}n_{j}^{l-1}\right)
\end{equation}
where $\alpha$ is a nonlinear activation function, and $W_1$ is a trainable weight matrix. $\Gamma_{ij}$ is the attention coefficient of hyperedge $e_j$ on node $v_{i}$ defined as

\begin{equation}
    \Gamma_{ij}= \frac{\exp({\mathbf{e_j}})}{\sum_{e_k\in E_i}\exp( \mathbf{e_k})}
\end{equation}

\begin{equation}
    \mathbf {e_{j}} = \beta {(W_{2}n_j^{l-1} * W_{3}p_i^{l-1})}
\end{equation}
where $E_i$ is the set of hyperedges $v_i$ is connected with, $\beta$ is a LeakyReLU activation function and $*$ is the element-wise multiplication and $W_{2}$, $W_{3}$ are trainable weights. 
\\
\textbf{Node-to-node level attention:} 
The hyperedge-to-node level attention captures global context information while generating node representations. However, while a subsequence is common in the different sequences, they may also have different roles and importance in each sequence. Therefore, it is crucial to capture the local information of nodes specific to hyperedges. Moreover, we need to incorporate  the individual contributions of adjacent local subsequences into the representation of a specific subsequence. Furthermore, retaining the positional information of the subsequences in a sequence is essential for the accurate analysis of sequence data. To address these, we introduce a node-to-node attention layer that passes information between nodes in a hyperedge. It learns the importance of subsequences for each other within the same sequence and also incorporates a position encoder that assigns a unique position to each subsequence. To get the position information, we adopt a simple positional encoder inspired by the transformer model~\cite{vaswani2017attention}. This enables us to preserve 
local and spatial information of a subsequence within a specific sequence. Using the attention mechanism, the $l$-th layer node feature $m_{i,j}^l$ of node $v_i$ belonging to hyperedge $e_j$ is defined as

\begin{equation}
m_{i,j}^{l}=\alpha \left(\sum_{v_{y}\in {e_{j}}}\Phi_{iy}W_{4}\bar{p_{y}}^{l}\right)
\end{equation}

\begin{equation}
\bar{p}_y = p_y + \text{{PE}}(pos_{v_y})
\end{equation}

\begin{equation}
\text{{PE}}(pos, 2x) = \sin(pos/10000^{2x/d})
\end{equation}

\begin{equation}
\text{{PE}}(pos, 2x+1) =\cos(pos/10000^{2x/d})
\end{equation}

Where $W_4$ is a trainable weight, $\Phi_{iy}$ is the attention coefficient of neighbor node $v_y$ on node $v_{i}$. $\text{PE}$ represents the positional encoding function, $pos_{v_y}$ represents the original positional index of $v_y$ in the sequence, $\text{PE}(pos, x)$ refers to the $x$-th dimension of the positional encoding of the word at position $pos$ in the sequence, and $d$ denotes the dimension of the positional encoding. The attention coefficient $\Phi_{iy}$ is defined as
\begin{equation}
    \Phi_{iy}= \frac{\exp({\mathbf{q_y}})}{\sum_{q_k\in e_j}\exp( \mathbf{q_k})}
\end{equation}

\begin{equation}
    \mathbf {q_{y}} = \beta {(W_{5}\bar{p_y}^{l} * W_{6}\bar{p_i}^{l})}
\end{equation}
where $e_j$ is the hyperedge node $v_i$ belongs, $W_{5}$ and $W_{6}$ are trainable weights. 
\\
\textbf{Node-to-hyperedge level attention:} Hyperedge is degree-free  consisting of an arbitrary number of nodes. But the contribution of nodes in hyperedge construction may not be the same. To highlight the important nodes, we employ an attention mechanism. This attention aggregates node representations and assigns higher weights to crucial ones. Moreover, during the aggregation process, it considers the representations of the nodes from both local and global contexts, allowing for a comprehensive understanding of their significance within the hypergraph. With the attention mechanism, $l$-th layer hyperedge feature $n_{j}^l$ of hyperedge $e_j$ is defined 

\begin{equation}
\label{eqn_n_7}
n_{j}^{l}=\alpha \left(\sum_{v_{i}\in {e_{j}}}\Delta_{ji}W_{7}(m_{i,j}^{l}||p_{i}^{l})\right)
\end{equation}
where $W_{7}$ is a trainable weight, || is the concatenation operation, and $\Delta_{ji}$ is the attention coefficient of node $v_i$ in the hyperedge $e_{j}$ defined as

\begin{equation}
    \Delta_{ji}= \frac{\exp({\mathbf{v_i}})}{\sum_{v_k\in e_j}\exp( \mathbf{v_k})}
\end{equation}

\begin{equation}
    \mathbf {v_{i}} = \beta  \left(W_{8}(m_{i,j}^{l} || p_i^{l}) * W_{9}n_{j}^{l-1})\right)
\end{equation}
where $v_{k}$ is the node that belongs to hyperedge $e_{j}$, and $W_{8}$, $W_{9}$ are trainable weights.

\sht\ generates the hyperedge representations by employing this three-level of attention. Finally, we linearly project the output of \sht\ with a trainable weight matrix to generate a $C$ dimensional output for each hyperedge as $ Z = nW_c^{T}$, where $C$ is the number of classes, $n$ is the output of the \sht, and $W_c$ is a trainable weight. We train our entire model using a cross-entropy loss function in a semi-supervised fashion.

\subsection{Complexity}
\sht\ is an efficient model that can be paralyzed across the edges and the nodes~\cite{velivckovic2017graph}. The time complexity of \sht\ can be expressed in terms of the cumulative complexity of the attention layers.  From equation \ref{eqn_n_4}, we can formulate the time complexity for the hyperedge-to-node level attention as: $O(|E|f{f}^{'}+|V|\kappa{f}^{'})$, where $\kappa$ is the average degree of nodes. And in the node-to-node level attention, the time complexity is: $O(|E|(\chi f{f}^{'}+\chi^{2}f^{'}))$, where $\chi$ is the average degree of hyperedges. Similarly, we can formulate the time complexity in the node-to-hyperedge level attention as:
 $O(|V|f{f}^{'}+|E|\chi{f}^{'})$.

\section {Experiment} \label{sec:experiment}
In this section, we perform extensive experiments on four different datasets and five different research problems to evaluate the proposed \sht\ model. We compute three different accuracy metrics, Precision (P), Recall (R), and F1-score (F1), to analyze and compare our proposed model with the state-of-the-art baseline models. This section starts with a description of our datasets, parameter settings, and baselines, and then we present our experimental results. 




\begin{table}[h!]
 \centering
    \caption{Number of Nodes (N) in the hypergraph for different datasets based on frequency threshold of ESPF and $k$ value of $k$-mer}\vspace{-2mm}
      \normalsize
  \begin{tabular}{|c|c|c|c|c|}
    \hline
    \cellcolor{gray!60} \textbf{ESPF} &
     \cellcolor{gray!60} \textbf{HD} &
     \cellcolor{gray!60} \textbf{BC} &
     \cellcolor{gray!60} \textbf{ACP} &
     \cellcolor{gray!60} \textbf{CPS} 
    
      \\
      & 
     $|N|$ &
     $|N|$ &
      $|N|$ &
      $|N|$
    \\
    \hline
    5 & 25207 & 446 &  382  & 10776  \\
    \hline
    10 & 15774 & 347 &225   &7987  \\
    \hline
    15 & 11166 &287 &166  &6769  \\
    \hline
    20 & 8871 & 253 & 138  & 5971  \\
    \hline
    25 & 7483 & 198 & 110  & 5477 \\
    \hline
  \end{tabular}
 \vspace{1mm}
 \centering
      \normalsize
  \begin{tabular}{|c|c|c|c|c|}
    \hline
    \cellcolor{gray!60} \textbf{$k$-mer} &
     \cellcolor{gray!60} \textbf{HD} &
     \cellcolor{gray!60} \textbf{BC} &
     \cellcolor{gray!60} \textbf{ACP} &
     \cellcolor{gray!60} \textbf{CPS} 
     
      \\
      & 
     $|N|$ &
     $|N|$ &
      $|N|$ &
      $|N|$
    \\
    \hline
    5 & 1247 & 3220 & 10301 &  99794 \\
    \hline
    10 & 602,855 & 14462 &6799 & 157,399 \\
    \hline
    15 & 1,449,240 &16163  &3103 & 193,544 \\
    \hline
    20 & 1,462,963 &17909 & -&  223,073  \\
    \hline
    25 &1,467,256 &18752 & -& 248,938  \\
    \hline
  \end{tabular}
  \label{tab1} \vspace{-5mm}
\end{table}

\subsection{Dataset}
We evaluate the performance of our model using four different sequence datasets. 
They are (1) \textbf{Human DNA} sequence, (2) \textbf{Anti-cancer peptides}, (3) \textbf{Cov-S-Protein-Seq} and (4) \textbf{Bach choral} harmony. All these datasets are publicly available online. 

1. \textbf{Human DNA} (HD) sequence dataset consists of 4,380 DNA sequences having seven different gene families \cite{DNA}.
2. The \textbf{Anti-cancer peptides} (ACPs) are short bioactive peptides \cite{Xie}. The ACP dataset contains 949 one-letter amino-acid sequences representing peptides and their four different anti-cancer activities.
3. The \textbf{Cov-S-Protein-Seq} (CPS) dataset consists of 1,238 spike protein sequences from 67 different coronavirus (CoV) species, including SARS-CoV-2 responsible for the COVID-19 pandemic \cite{Sarwan, Sandrine}. The dataset provides information on the seven different CoV species (CVS) and their six different host species (CHS) \cite{Kiril}.
4. Music is sequences of sounding events. Each event has a specific chord label. The \textbf{Bach choral} (BC) harmony consists of 5,667 events of five different cord labels \cite{UCI}. Appendix \ref{dataset_description} provides a more detailed description of the dataset.

\vspace{-3mm}
\subsection{Parameter Settings}
We extract subsequences from given sequence datasets using ESPF and $k$-mer separately to create our hypergraph. 
With a low-frequency threshold, ESPF produces more subsequences, and all of them may not be important. But with a large-frequency threshold, it produces less number of subsequences and there might be a missing of some vital subsequences. We choose five different frequency thresholds from 5 to 25 and examine the impact of threshold value on hypergraph learning. Similarly, in $k$-mer, typically with the increment of $k$-mer length (i.e., $k$ value), the number of subsequences also increases. We choose five different $k$ values from 5 to 25 and examine their impact on hypergraph learning. However, as Anticancer peptide sequences are too small, we just choose $k$ from 5 to 15.  The number of nodes for different threshold values of ESPF and $k$ value of $k$-mer is given in Table~\ref{tab1} for each dataset.
We perform a random split of our datasets, dividing them into 80\% for training, 10\% for validation, and 10\% for testing. This splitting process is repeated five times, and the average accuracy metrics are calculated and reported in the results section. To find the optimal hyperparameters, a grid search method is used on the validation set. The optimal learning rate is determined to be 0.001, and the optimal dropout rate is found to be 0.3 to prevent overfitting.

We utilize a single-layer \sht\ having a three-level of attention network. Simple one-hot coding is used as an initial feature of the sequences. A LeakyReLU activation function is used on the attention networks side. The model is trained with 1000 epochs and optimized using Adam optimizer. An early stop mechanism is used if the validation accuracy does not change for 100 consecutive epochs. 

The ML classifiers in subsection: \ref{sec:baselines} are taken from sci-kit learn \cite{scikit-learn}. For logistic regression (LR), we set the inverse of regularization strength, C=2. A linear kernel with polynomial degree 3 is used in the support vector machine (SVM). Default parameters are used for the decision tree (DT) classifier.
The DL models are implemented from Keras layers \cite{chollet2015keras}. In the DL models, RCNN and BiLSTM, we use relu and softmax activation functions in the hidden dense layers and dense output layer, respectively. The models are optimized using Adam optimizer, and dropout layers of 0.3 are used. For node2vec, the walk length, number of walks, and window size are set to 80, 10, and 10, respectively, and for graph2vec, we use the default parameters following the source. We follow DGL \cite{Minjie} 
to implement graph attention network (GAT).  For DNA-GCN, and hypergraph neural networks (HGNN, HyperGAT), we use the same hyper-parameters as mentioned in the source papers.

\subsection{Baselines}\label{sec:baselines} 
We evaluate our model by comparing it with different types of state-of-the-art models. The baseline models are categorized into:
\textit{Machine Learning:} CountVectorizer is used to generate input features, with logistic regression (LR), support vector machine (SVM), and decision tree (DT) classifiers applied. \textit{Deep Learning:} Recurrent convolutional neural networks (RCNN) and bidirectional long short-term memory (BiLSTM) are used. \textit{Node2vec:} We create De-Bruijn graph from each sequence as explained in section~\ref{sec:works}. Then we apply Node2vec \cite{grover2016node2vec}, which is a random walk-based graph embedding that generates the embedding of nodes. To obtain the graph-level representation, we average the embedding of nodes of that graph and feed it as a feature to the ML classifier for sequence classification. \textit{Graph2vec:} The Graph2vec \cite{narayanan2017graph2vec} method is applied to the De-Bruijn graph to generate graph embeddings, which are fed into machine learning classifiers for sequence classification. \textit{Graph Neural Network:} graph attention network (GAT) \cite{velivckovic2017graph} is applied on De-Bruijn graphs to learn node embedding. Graph-level embedding is obtained using an average pooling-based readout function. Moreover, we follow DNA-GCN \cite{Guo} to construct a heterogeneous graph from the entire corpus and the extracted subsequences. After constructing the graph, we apply a two-layer GCN. \textit{Hypergraph Neural Network (HNN):} Model performances are compared with two state-of-the-art hypergraph neural network models: HGNN~\cite{Feng2018} and HyperGAT~\cite{ding2020more}. HGNN generates node representations by aggregating hyperedges, which are then combined to get hyperedge representations. HyperGAT, an attention-based hypergraph neural network, generates embeddings of subsequences, which are then pooled to get sequence embedding. In both HGNN and HyperGAT, one-hot coding of nodes is used as the initial feature.

\vspace{-2mm}
\subsection{Results}

\subsubsection {Model Performance}
\begin{figure}[h]
    \centering{ 
    \includegraphics[width= 0.47\textwidth]{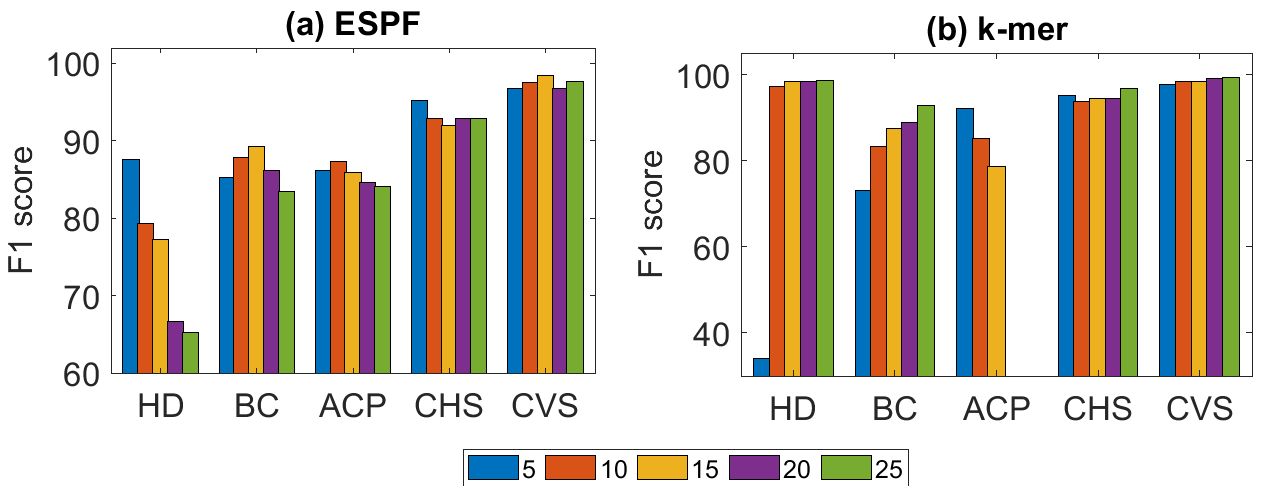}\vspace{-4mm}
   \caption{Performance comparison of the proposed model with different thresholds of (a) ESPF and (b) $k$-mer for different datasets.}
    \label{fig:result_graph1}} \vspace{-4mm}
\end{figure}

We evaluate our proposed model using extensive experiments on four different datasets across five problems with different threshold values of ESPF and $k$-mer. Figure \ref{fig:result_graph1} shows the performance of our models in terms of the F1-score. 

\begin{table*}[t]
\centering
\caption{ \centering{Performance comparisons of models for Human DNA, Bach choral and Anticancer datasets } }
\label{table:compsres}
\begin{tabular}{|c|c| c c c | c c c | c c c|} 
 \hline
 
 \multirow{2}{*}
 { Model }
 &{Method}
 & \multicolumn{3}{c|}{\cellcolor{gray!60}\textbf{Human DNA}}
 & \multicolumn{3}{c|}{\cellcolor{gray!60}\textbf{Bach choral}}
 & \multicolumn{3}{c|}{\cellcolor{gray!60}\textbf{Anticancer pept.}} 
 \\

 & & \cellcolor{gray!25}\textbf{P} & \cellcolor{gray!25}\textbf{R} & \cellcolor{gray!25}\textbf{F1} 
 
 & \cellcolor{gray!25}\textbf{P} & \cellcolor{gray!25}\textbf{R} & \cellcolor{gray!25}\textbf{F1}
 
 & \cellcolor{gray!25}\textbf{P} & \cellcolor{gray!25}\textbf{R} & \cellcolor{gray!25}\textbf{F1}
 
 \\ \hline
 
 & LR  & 92.82 & 90.64 & 90.84 & 88.09 & 76.68 & 78.52 & 77.00 & 83.16 & 77.67 \\
 {ML} & SVM & 90.09 & 85.39 & 85.83 & 89.30 & 80.27 & 82.08 & 83.10 & 86.32 & 83.27 \\
   & DT & 92.87 & 80.37 & 83.68 & 86.49 & 70.85 & 73.92 & 78.28 & 85.32 & 81.35 \\

\hline

& RCNN & 68.84 & 37.90 & 27.86 & 76.83 & 71.30 & 68.23 & 69.62 & 80.00 & 73.34 \\
 {DL} & BiLSTM & 77.80 & 39.27 & 35.18 & 73.93 & 69.96 & 66.32 & 65.70 & 81.05 & 72.57\\
 \hline
 
& LR & 36.09 & 32.19 & 22.89 & 16.11 & 20.17 & 18.14 & 71.32 & 82.11 & 75.11\\
{Node2vec} & SVM & 10.32 & 30.82 & 14.52 & 14.14 & 20.18 & 16.17 & 66.39 & 81.05 & 72.99\\
& DT & 18.04 & 18.26 & 18.13 & 23.03 & 22.87 & 22.89 & 68.75 & 64.21 & 66.40 \\
\hline

 & LR & 21.07 & 26.94 & 23.63 & 23.34 & 19.73 & 17.81 & 66.39 & 81.05 & 72.99\\
 {Graph2vec} & SVM & 10.32 & 30.82 & 14.52 & 13.09 & 15.75 & 16.50 & 71.32 & 82.11 & 75.11\\
& DT & 19.41 & 19.63 & 19.39 & 25.60 & 25.56 & 25.48 & 77.93 & 73.68 & 75.54 \\

\hline

{GNN} & DNA-GCN & 96.46 & 96.28 &96.36 & 85.54 & 85.24 & 85.27 & 83.25 & 83.53 & 83.82 \\
& GAT & 30.06 & 42.14 & 36.01 & 24.75 & 29.19 & 31.12 & 79.23 & 87.44 & 79.67 \\
\hline

{HNN} & HGNN & 87.03 & 86.82 &87.12 & 86.12 & 86.89 & 86.93 & 83.82 & 85.42 & 83.97 \\
& HyperGAT & 85.13 & 85.33 & 84.11 & 88.09 & 87.44 & 87.45 & 85.33 & 88.42 & 86.68 \\

\hline

 & ESPF & 88.77 & 87.89 & 87.78 & 89.93 & 89.72 & 89.88 & 91.98 & 86.75 & 87.65 \\
 \texttt{\sht} & \textbf{$k$-mer} & \textbf {98.91} & \textbf{98.88} & \textbf{98.83} &  \textbf {93.78} & \textbf{93.10} & \textbf{93.18} & \textbf {93.36} & \textbf{91.72} & \textbf{92.33} \\
\hline
\end{tabular}
\label{tab3}\vspace{-1mm}
\end{table*}
Figure \ref{fig:result_graph1} (a) illustrates the performance of our models on different datasets as the ESPF frequency threshold varies from 5 to 25. Generally, an increase in the ESPF frequency threshold corresponds to a decline in model performance. This impact is more noticeable for the Human DNA dataset, where a change in the frequency threshold from 5 to 25 leads to a nearly 25\% reduction in the F1 score. This could be due to a decrease in the number of subsequences (nodes) at higher frequency thresholds, potentially impacting the learning of hyperedges. Optimal performance is achieved at different frequency thresholds for different datasets: 5 for Human DNA and CoV-S-Protein-Seq (Host species classification), 15 for Bach Choral and CoV-S-Protein-Seq (CoV species classification), and 10 for Anticancer peptides.
Figure \ref{fig:result_graph1} (b) displays the performance of our models on different datasets as the $k$ value in $k$-mer changes from 5 to 25. The Human DNA dataset shows the most substantial change in response to different $k$ values, with the F1-score rising about 190\% from 33.98\% at $k=5$ to 98.83\% at $k=25$. A noticeable increase in F1-score (27\%) is also seen for the Bach choral dataset, with $k$ rising from 5 to 25. The value of $k$ has less impact on the Cov-S-Protein-Seq datasets, while performance decreases for the Anticancer peptides dataset as $k$ increases. This can be attributed to the changing number of nodes, as seen in Table \ref{tab1}: as $k$ increases, the number of nodes significantly grows in the Human DNA dataset, improving performance, whereas the decreasing node count for Anticancer peptides with growing $k$ may negatively affect performance.
\vspace{-1.0mm}
\subsubsection {Comparative analysis with Baselines} We compare our model with various state-of-the-art baselines for each dataset, using the best ESPF and $k$-mer thresholds determined for our models. Table \ref{tab3} shows results for the Human DNA, Bach choral, and Anticancer peptides datasets, while Table~\ref{tab4} presents results for the Cov-S-Protein-Seq dataset. Our models consistently outperform the baselines across all datasets. Specifically, for the Human DNA dataset, our \sht\ model with $k$-mer yields the best performance (98.91\% Precision, 98.88\% Recall, and 98.83\% F1-score), significantly higher than the next best model, DNA-GCN. Moreover, ML models deliver competitive results with over 80\% F1-score.

Eventually, for all the datasets, the hypergraph-based model gives the best performance. Specifically, in almost every case, HyperGAT serves as the superior baseline, while HGNN performs as the second-best baseline. The reason for their success lies in their ability to capture higher-order, intricate relationships within a hypergraph structure.  Additionally, HyperGAT utilizes attention networks to enhance sequence representation learning, which is superior to HGNN's approach. It is worth mentioning that we apply these models to our hypergraphs, and our experiments demonstrate the effectiveness of representing sequences as hypergraphs.

\begin{table*}[t]
\centering
\caption{ \centering{Performance comparisons of models on Cov-S-Protein-Seq dataset for Host and CoV species prediction } }
\label{table:compsres}
\begin{tabular}{|c|c| c c c | c c c | }
 \hline
 \multirow{2}{*}
 { Method }
 &{Model}
 & \multicolumn{3}{c|}{\cellcolor{gray!60}\textbf{ Host species}}
 & \multicolumn{3}{c|}{\cellcolor{gray!60}\textbf{ CoV species}}
 \\
 
 & & \cellcolor{gray!25}\textbf{P} & \cellcolor{gray!25}\textbf{R} & \cellcolor{gray!25}\textbf{F1} 
 
 & \cellcolor{gray!25}\textbf{P} & \cellcolor{gray!25}\textbf{R} & \cellcolor{gray!25}\textbf{F1}
 
 \\ \hline
 
 & LR  & 92.64 & 91.94 & 91.48 & 96.04& 95.16 & 95.21 \\
 {ML} & SVM & 94.20 & 93.55 & 93.42 & 96.60 & 95.97 & 96.02 \\
   & DT & 92.25 & 91.13 & 91.25 & 95.60 & 94.97 & 95.02 \\

\hline

& RCNN & 83.22 & 79.03 & 76.82 & 65.53 & 62.90 & 57.47 \\
 {DL} & BiLSTM & 70.61 & 66.94 & 61.56 & 66.66 & 72.58 & 66.92 \\
 
\hline
 & LR &26.28 & 29.03 & 19.29 &12.34 & 20.16 & 14.92\\
 {Node2vec} & SVM & 23.07 & 24.19 & 23.27 & 22.42 & 25.00 & 23.54 \\
  & DT & 20.00 & 21.77 & 20.80 & 23.47 & 23.39 & 23.33 \\
  
\hline

 & LR & 23.74 & 25.00 &23.98 & 25.92 & 28.23 & 26.67\\
 {Graph2vec} & SVM & 17.22& 27.42 &17.81 & 17.16 & 22.58 & 16.59 \\
& DT & 22.68 & 22.58 & 22.50 & 22.80 & 22.58 & 22.47 \\
\hline

{GNN} & DNA-GCN &90.91 & 90.18 & 91.11 & 94.34 & 94.57 & 94.13 \\
& GAT & 22.36 & 33.23 & 25.22 & 24.10 & 29.65 & 26.19 \\
\hline

{HNN} & HGNN & 91.52 & 91.91 &91.60 & 94.62 & 94.42 & 94.63 \\
& HyperGAT & 93.44 & 93.55 & 93.14 & 95.52 & 95.35 & 95.45 \\
\hline

 & ESPF & 95.78 & 95.66 & 95.49 & 98.89 & 98.78 & 98.72 \\
 \texttt{\sht} & \textbf{$k$-mer} & \textbf {97.83} & \textbf{96.13} & \textbf{97.01} &  \textbf {99.56} & \textbf{99.29} & \textbf{99.45} \\
\hline
\end{tabular}
\label{tab4} \vspace{-2.5mm}
\end{table*}

DNA-GCN demonstrates competitive performance due to its heterogenous graph-based structure, allowing information passage between sequences and subsequences. Among ML models, SVM generally outperforms others.  Notably, the DL models perform less effectively than ML models across all datasets. For instance, in the Bach choral dataset, \sht\ with $k$-mer achieved over 93\% F1-score, while SVM achieved over 82\%. In contrast, DL models like RCNN and BiLSTM only reached 68.23\% and 66.32\% F1-score respectively. This could be due to DL models being data-hungry and our datasets being relatively small, limiting their performance.


Table \ref{tab3} and \ref{tab4} reveal a disparity in the performance of Node2vec, Graph2vec, and GAT across datasets. While their performance on the Anticancer peptides dataset is impressive, it is considerably lower for the other datasets. This variance could be due to differences in the network structures of these datasets. Upon investigating, we find that the average network density of each dataset could be a contributing factor. The Anticancer peptides dataset, for instance, has the highest average network density of 0.3868, while the Human DNA, Bach Choral, and CoV-S-Protein-Seq dataset graphs have lower densities of 0.0100, 0.2702, and 0.0031, respectively. This difference in network density might be influencing the performance of Node2vec, Graph2vec, and GAT.

In brief, \sht\ with $k$-mer delivers better performances than \sht\ with ESPF. This may be because ESPF assumes frequent subsequences are the only important ones and eliminates many infrequent subsequences. However, some infrequent subsequences may also be important. Thus using ESPF, we may seldom lose some infrequent important ones. On the contrary, $k$-mer does not lose any subsequences; rather it fetches all and lets the attention model discover the critical ones. In general, a larger $k$-mer is preferable since it provides greater uniqueness and helps to eliminate the repetitive substrings. 

\begin{figure}[h]
    \centering\includegraphics[width=0.45\textwidth, height=7.5cm]{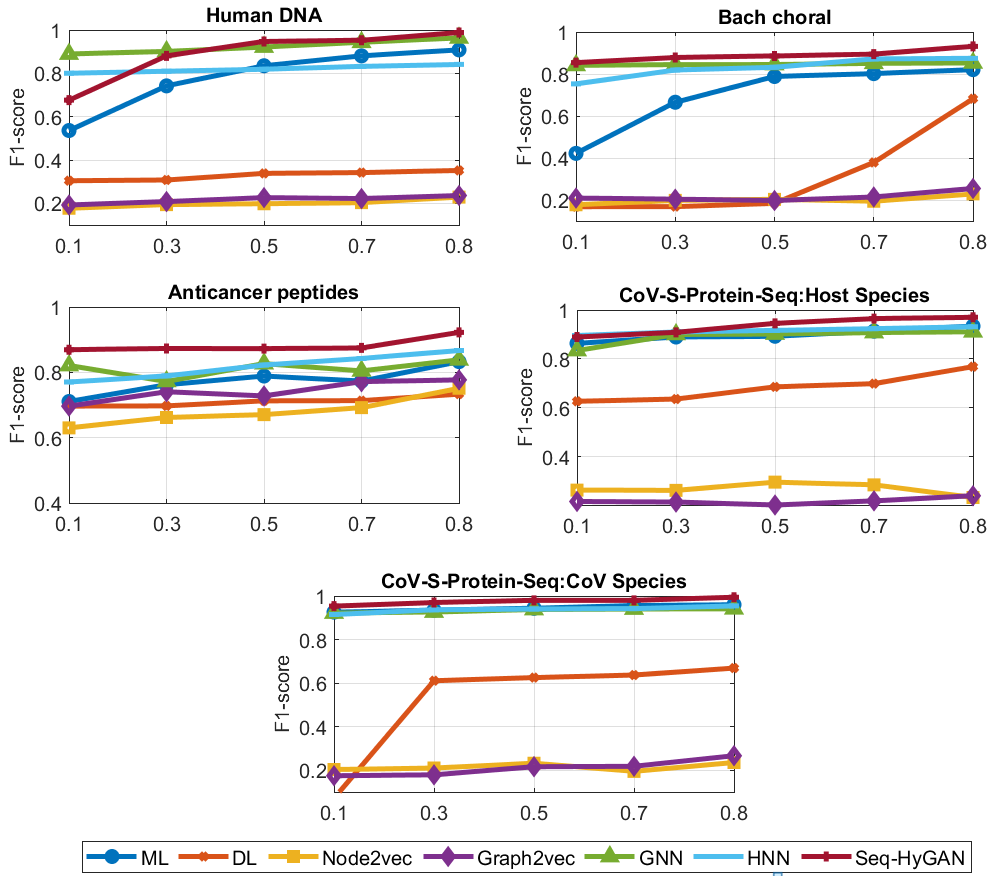}\vspace{-2mm}
   \caption{Performance comparison of models for different training sizes}
    \label{fig:result_graph}\vspace{-4mm}
\end{figure}

Moreover, we select the top-performing method from each baseline model for each dataset and compare their performance with our model as we vary the training data sizes from 10\% to 80\%. In this comparison, shown in Fig \ref{fig:result_graph} in terms of F1-score, our model consistently outperforms the others, even with limited training data. However, some baseline models significantly underperform with decreased training sizes.


The hypergraph's innate ability to capture complex higher-order relationships has made it an effective model for many scientific studies. \sht\ leverages a hypergraph structure and captures higher-order intricate relations of subsequences within a sequence and between the sequences. Furthermore, it generates a much more robust representation of sequence by utilizing an attention mechanism that discovers the important subsequences of that sequence. While GAT \cite{velivckovic2017graph} also uses an attention mechanism, it is limited to learning important neighbors of a node and cannot learn significant edges. Additionally, GAT is unsuitable for complex networks with triadic or tetradic relations. The key strength of our proposed model, \sht, lies in its three-level attention mechanism, which effectively captures both local and global information. This mechanism allows for the generation of node representations by aggregating information from connected hyperedges (global information) and neighboring nodes (local information) within the same hyperedge, with a specific emphasis on important ones. Likewise, it enables the generation of hyperedge representations by aggregating member nodes, with a particular focus on critical ones.

\begin{table}[t]
\centering
\caption{ \centering{F1-score for different varients} }\vspace{-3mm}
\label{table:compsres}
\begin{tabular}{|c| c c | c c | }
 \hline
 \multirow{2}{*}
 { Dataset }
 & \multicolumn{2}{c|}{\cellcolor{gray!60}\textbf{Line graph }}
 & \multicolumn{2}{c|}{\cellcolor{gray!60}\textbf{ \sht}}
 \\
 
 &  \cellcolor{gray!25}\textbf{GCN} & \cellcolor{gray!25}\textbf{GAT} 
 
 & \cellcolor{gray!25}\textbf{w/o attn} & \cellcolor{gray!25}\textbf{w/ attn} 
 
 \\ \hline
 
 {Human DNA} & 37.08 & 39.19 & 94.13 & 98.83\\
\hline
 {Bach coral} &  75.41 & 77.65 & 91.21 & 93.18 \\
\hline
 {Anticancer peptides} & 83.52 & 86.86 & 89.88 & 92.33  \\
\hline
 {Host species} & 72.70 & 75.31 & 90.16 &97.01 \\
\hline
{CoV species} &80.83 & 86.50 & 95.77 & 99.45 \\
\hline
\end{tabular}
\label{tab5} \vspace{-4.5mm}
\end{table} 

\subsubsection{Space Analysis}
Our models demonstrate efficient memory usage by creating only one hypergraph per dataset. Regardless of the chosen thresholds for ESPF and $k$ for $k$-mer, the number of hyperedges remains consistent across the hypergraphs. In contrast, the De-Bruijn method constructs a separate graph for each sequence, resulting in a significant number of nodes and edges. For instance, the hypergraph constructed from the Human DNA dataset with a $k$-mer value of 25 contains 1,467,256 nodes and 4380 hyperedges. In comparison, the De-Bruijn graph constructed from the same dataset with the same $k$-mer value comprises 5,422,447 nodes and 5,418,151 edges. Similar trends are observed in other datasets as well, which can be found in Appendix \ref{node_edge_de_bruijn}.

\subsubsection{Case Study - Impact of hypergraph structure}
In contrast to standard graphs, hypergraphs offer the ability to capture higher-order complex relationships that are not easily represented by standard graphs. To demonstrate this capability, we conduct a comparison between our hypergraph-based \sht\ model and standard graphs. To facilitate this comparison, we construct line graphs from the same datasets, where each sequence is represented as a node, and nodes are connected if they share a certain number ($S$) of common subsequences (with $S=2$ in our case). Subsequently, we apply GCN and GAT independently to learn node representations and classify sequences. The performance results are presented in Table \ref{tab5}. The results clearly indicate that our hypergraph-based \sht\ models outperform line graph GCN and GAT models in terms of the F1-score.

\vspace{-2mm}
\subsubsection{Case Study - Impact of Attention Network}
In this research paper, we aim to investigate the impact of the attention network in the proposed \sht\ model on classification performance. To achieve this, we train the model separately with and without the attention network on all datasets and classification problems. The corresponding F1-scores for the test datasets were recorded and are presented in Table \ref{tab5}. The results show that the proposed \sht\ model with the attention network (w/ attn) outperforms the model without the attention network (w/o attn). Specifically, for CoV-S-Protein-Seq: Host species, the F1-score improved from 90.16\% without the attention network to 97.01\% with the attention network, representing a significant 7.59\% improvement. This improvement can be attributed to the attention network's ability to discover crucial subsequences while learning the sequence representation, which ultimately leads to better performance.
\vspace{-1mm}
\section{Conclusion}\label{sec:con}
In this paper, we introduce a novel Sequence Hypergraph Attention Network for sequence classification. Unlike previous models, we leverage a unique hypergraph structure to capture complex higher-order structural similarities among sequences. While sequences are represented as hyperedge, subsequences are represented as nodes in the hypergraph. With our three-level attention model, we learn hyperedge representations as sequences while considering the importance of sequence and subsequences for each other. Our extensive experiments demonstrate that our method surpasses various baseline models in terms of performance. we also show that our model is space and time efficient with one compact hypergraph setting.
\vspace{-2mm}

\begin{acks}
 This work has been supported partially by the National Science Foundation under Grant No 2104720.
\end{acks}

\bibliographystyle{ACM-Reference-Format}
\bibliography{MLG}


\begin{thebibliography}{47}


\ifx \showCODEN    \undefined \def \showCODEN     #1{\unskip}     \fi
\ifx \showDOI      \undefined \def \showDOI       #1{#1}\fi
\ifx \showISBNx    \undefined \def \showISBNx     #1{\unskip}     \fi
\ifx \showISBNxiii \undefined \def \showISBNxiii  #1{\unskip}     \fi
\ifx \showISSN     \undefined \def \showISSN      #1{\unskip}     \fi
\ifx \showLCCN     \undefined \def \showLCCN      #1{\unskip}     \fi
\ifx \shownote     \undefined \def \shownote      #1{#1}          \fi
\ifx \showarticletitle \undefined \def \showarticletitle #1{#1}   \fi
\ifx \showURL      \undefined \def \showURL       {\relax}        \fi
\providecommand\bibfield[2]{#2}
\providecommand\bibinfo[2]{#2}
\providecommand\natexlab[1]{#1}
\providecommand\showeprint[2][]{arXiv:#2}

\bibitem[Aktas and Akbas(2021)]%
        {Aktas2021}
\bibfield{author}{\bibinfo{person}{Mehmet~Emin Aktas} {and}
  \bibinfo{person}{Esra Akbas}.} \bibinfo{year}{2021}\natexlab{}.
\newblock \showarticletitle{Hypergraph Laplacians in Diffusion Framework}.
\newblock \bibinfo{journal}{\emph{Studies in Computational Intelligence}}
  \bibinfo{volume}{1016} (\bibinfo{date}{2} \bibinfo{year}{2021}),
  \bibinfo{pages}{277--288}.
\newblock
\showISBNx{9783030934125}
\showISSN{18609503}
\urldef\tempurl%
\url{https://doi.org/10.48550/arxiv.2102.08867}
\showDOI{\tempurl}


\bibitem[Aktas et~al\mbox{.}(2021)]%
        {Aktas2021a}
\bibfield{author}{\bibinfo{person}{Mehmet~Emin Aktas}, \bibinfo{person}{Thu
  Nguyen}, \bibinfo{person}{Sidra Jawaid}, \bibinfo{person}{Rakin Riza}, {and}
  \bibinfo{person}{Esra Akbas}.} \bibinfo{year}{2021}\natexlab{}.
\newblock \showarticletitle{Identifying critical higher-order interactions in
  complex networks}.
\newblock \bibinfo{journal}{\emph{Scientific Reports 2021 11:1}}
  \bibinfo{volume}{11} (\bibinfo{date}{10} \bibinfo{year}{2021}),
  \bibinfo{pages}{1--11}.
\newblock
Issue 1.
\showISBNx{0123456789}
\showISSN{2045-2322}
\urldef\tempurl%
\url{https://doi.org/10.1038/s41598-021-00017-y}
\showDOI{\tempurl}


\bibitem[Ali et~al\mbox{.}(2022)]%
        {Sarwan}
\bibfield{author}{\bibinfo{person}{Sarwan Ali}, \bibinfo{person}{Babatunde
  Bello}, \bibinfo{person}{Prakash Chourasia}, \bibinfo{person}{Ria~Thazhe
  Punathil}, \bibinfo{person}{Yijing Zhou}, {and} \bibinfo{person}{Murray
  Patterson}.} \bibinfo{year}{2022}\natexlab{}.
\newblock \showarticletitle{PWM2Vec: An Efficient Embedding Approach for Viral
  Host Specification from Coronavirus Spike Sequences}.
\newblock \bibinfo{journal}{\emph{Biology}}  \bibinfo{volume}{11}
  (\bibinfo{date}{3} \bibinfo{year}{2022}).
\newblock
Issue 3.
\showISSN{2079-7737}
\urldef\tempurl%
\url{https://doi.org/10.3390/BIOLOGY11030418}
\showDOI{\tempurl}


\bibitem[Ashoor et~al\mbox{.}(2020)]%
        {Ashoor2020}
\bibfield{author}{\bibinfo{person}{Haitham Ashoor}, \bibinfo{person}{Xiaowen
  Chen}, \bibinfo{person}{Wojciech Rosikiewicz}, \bibinfo{person}{Jiahui Wang},
  \bibinfo{person}{Albert Cheng}, \bibinfo{person}{Ping Wang},
  \bibinfo{person}{Yijun Ruan}, {and} \bibinfo{person}{Sheng Li}.}
  \bibinfo{year}{2020}\natexlab{}.
\newblock \showarticletitle{Graph embedding and unsupervised learning predict
  genomic sub-compartments from HiC chromatin interaction data}.
\newblock \bibinfo{journal}{\emph{Nature Communications 2020 11:1}}
  \bibinfo{volume}{11} (\bibinfo{date}{3} \bibinfo{year}{2020}),
  \bibinfo{pages}{1--11}.
\newblock
Issue 1.
\showISSN{2041-1723}
\urldef\tempurl%
\url{https://doi.org/10.1038/s41467-020-14974-x}
\showDOI{\tempurl}


\bibitem[Atkinson et~al\mbox{.}(2009)]%
        {atkinson2009using}
\bibfield{author}{\bibinfo{person}{Holly~J Atkinson}, \bibinfo{person}{John~H
  Morris}, \bibinfo{person}{Thomas~E Ferrin}, {and} \bibinfo{person}{Patricia~C
  Babbitt}.} \bibinfo{year}{2009}\natexlab{}.
\newblock \showarticletitle{Using sequence similarity networks for
  visualization of relationships across diverse protein superfamilies}.
\newblock \bibinfo{journal}{\emph{PloS one}} \bibinfo{volume}{4},
  \bibinfo{number}{2} (\bibinfo{year}{2009}), \bibinfo{pages}{e4345}.
\newblock


\bibitem[Belouzard et~al\mbox{.}(2012)]%
        {Sandrine}
\bibfield{author}{\bibinfo{person}{Sandrine Belouzard},
  \bibinfo{person}{Jean~K. Millet}, \bibinfo{person}{Beth~N. Licitra}, {and}
  \bibinfo{person}{Gary~R. Whittaker}.} \bibinfo{year}{2012}\natexlab{}.
\newblock \showarticletitle{Mechanisms of coronavirus cell entry mediated by
  the viral spike protein}.
\newblock \bibinfo{journal}{\emph{Viruses}}  \bibinfo{volume}{4}
  (\bibinfo{year}{2012}), \bibinfo{pages}{1011--1033}.
\newblock
Issue 6.
\showISSN{1999-4915}
\urldef\tempurl%
\url{https://doi.org/10.3390/V4061011}
\showDOI{\tempurl}


\bibitem[Bretto({[n.\,d.]})]%
        {Bretto}
\bibfield{author}{\bibinfo{person}{Alain Bretto}.}
  \bibinfo{year}{[n.\,d.]}\natexlab{}.
\newblock \showarticletitle{Hypergraph Theory}.
\newblock  (\bibinfo{year}{[n.\,d.]}).
\newblock
\urldef\tempurl%
\url{http://www.springer.com/series/8445}
\showURL{%
\tempurl}


\bibitem[Bumgardner et~al\mbox{.}(2022)]%
        {Bumgardner2022}
\bibfield{author}{\bibinfo{person}{Bri Bumgardner}, \bibinfo{person}{Farhan
  Tanvir}, \bibinfo{person}{Khaled~Mohammed Saifuddin}, {and}
  \bibinfo{person}{Esra Akbas}.} \bibinfo{year}{2022}\natexlab{}.
\newblock \showarticletitle{Drug-Drug Interaction Prediction: a Purely SMILES
  Based Approach}.
\newblock  (\bibinfo{date}{1} \bibinfo{year}{2022}),
  \bibinfo{pages}{5571--5579}.
\newblock
\showISBNx{9781665439022}
\urldef\tempurl%
\url{https://doi.org/10.1109/BIGDATA52589.2021.9671766}
\showDOI{\tempurl}


\bibitem[Chauhan({[n.\,d.]})]%
        {DNA}
\bibfield{author}{\bibinfo{person}{Nagesh~Singh Chauhan}.}
  \bibinfo{year}{[n.\,d.]}\natexlab{}.
\newblock \bibinfo{title}{Demystify DNA Sequencing with Machine Learning |
  Kaggle}.
\newblock
\newblock
\urldef\tempurl%
\url{https://www.kaggle.com/code/nageshsingh/demystify-dna-sequencing-with-machine-learning/notebook}
\showURL{%
\tempurl}


\bibitem[Chollet et~al\mbox{.}(2015)]%
        {chollet2015keras}
\bibfield{author}{\bibinfo{person}{Francois Chollet} {et~al\mbox{.}}}
  \bibinfo{year}{2015}\natexlab{}.
\newblock \bibinfo{booktitle}{\emph{Keras}}.
\newblock
\urldef\tempurl%
\url{https://github.com/fchollet/keras}
\showURL{%
\tempurl}


\bibitem[Compeau et~al\mbox{.}(2011)]%
        {compeau2011bruijn}
\bibfield{author}{\bibinfo{person}{Phillip~EC Compeau},
  \bibinfo{person}{Pavel~A Pevzner}, {and} \bibinfo{person}{Glenn Tesler}.}
  \bibinfo{year}{2011}\natexlab{}.
\newblock \showarticletitle{Why are de Bruijn graphs useful for genome
  assembly?}
\newblock \bibinfo{journal}{\emph{Nature biotechnology}} \bibinfo{volume}{29},
  \bibinfo{number}{11} (\bibinfo{year}{2011}), \bibinfo{pages}{987}.
\newblock


\bibitem[Ding et~al\mbox{.}(2020)]%
        {ding2020more}
\bibfield{author}{\bibinfo{person}{Kaize Ding}, \bibinfo{person}{Jianling
  Wang}, \bibinfo{person}{Jundong Li}, \bibinfo{person}{Dingcheng Li}, {and}
  \bibinfo{person}{Huan Liu}.} \bibinfo{year}{2020}\natexlab{}.
\newblock \showarticletitle{Be more with less: Hypergraph attention networks
  for inductive text classification}.
\newblock \bibinfo{journal}{\emph{arXiv preprint arXiv:2011.00387}}
  (\bibinfo{year}{2020}).
\newblock


\bibitem[Eickholt and Cheng(2013)]%
        {Eickholt2013}
\bibfield{author}{\bibinfo{person}{Jesse Eickholt} {and}
  \bibinfo{person}{Jianlin Cheng}.} \bibinfo{year}{2013}\natexlab{}.
\newblock \showarticletitle{DNdisorder: Predicting protein disorder using
  boosting and deep networks}.
\newblock \bibinfo{journal}{\emph{BMC Bioinformatics}}  \bibinfo{volume}{14}
  (\bibinfo{date}{3} \bibinfo{year}{2013}), \bibinfo{pages}{1--10}.
\newblock
Issue 1.
\showISSN{14712105}
\urldef\tempurl%
\url{https://doi.org/10.1186/1471-2105-14-88/FIGURES/6}
\showDOI{\tempurl}


\bibitem[Feng et~al\mbox{.}(2018)]%
        {Feng2018}
\bibfield{author}{\bibinfo{person}{Yifan Feng}, \bibinfo{person}{Haoxuan You},
  \bibinfo{person}{Zizhao Zhang}, \bibinfo{person}{Rongrong Ji}, {and}
  \bibinfo{person}{Yue Gao}.} \bibinfo{year}{2018}\natexlab{}.
\newblock \showarticletitle{Hypergraph Neural Networks}.
\newblock \bibinfo{journal}{\emph{33rd AAAI Conference on Artificial
  Intelligence, AAAI 2019, 31st Innovative Applications of Artificial
  Intelligence Conference, IAAI 2019 and the 9th AAAI Symposium on Educational
  Advances in Artificial Intelligence, EAAI 2019}} (\bibinfo{date}{9}
  \bibinfo{year}{2018}), \bibinfo{pages}{3558--3565}.
\newblock
\showISBNx{9781577358091}
\showISSN{2159-5399}
\urldef\tempurl%
\url{https://doi.org/10.48550/arxiv.1809.09401}
\showDOI{\tempurl}


\bibitem[Gligorijević et~al\mbox{.}(2021)]%
        {Gligorijevic2021}
\bibfield{author}{\bibinfo{person}{Vladimir Gligorijević},
  \bibinfo{person}{P.~Douglas Renfrew}, \bibinfo{person}{Tomasz Kosciolek},
  \bibinfo{person}{Julia~Koehler Leman}, \bibinfo{person}{Daniel Berenberg},
  \bibinfo{person}{Tommi Vatanen}, \bibinfo{person}{Chris Chandler},
  \bibinfo{person}{Bryn~C. Taylor}, \bibinfo{person}{Ian~M. Fisk},
  \bibinfo{person}{Hera Vlamakis}, \bibinfo{person}{Ramnik~J. Xavier},
  \bibinfo{person}{Rob Knight}, \bibinfo{person}{Kyunghyun Cho}, {and}
  \bibinfo{person}{Richard Bonneau}.} \bibinfo{year}{2021}\natexlab{}.
\newblock \showarticletitle{Structure-based protein function prediction using
  graph convolutional networks}.
\newblock \bibinfo{journal}{\emph{Nature Communications 2021 12:1}}
  \bibinfo{volume}{12} (\bibinfo{date}{5} \bibinfo{year}{2021}),
  \bibinfo{pages}{1--14}.
\newblock
Issue 1.
\showISSN{2041-1723}
\urldef\tempurl%
\url{https://doi.org/10.1038/s41467-021-23303-9}
\showDOI{\tempurl}


\bibitem[Grisoni et~al\mbox{.}({[n.\,d.]})]%
        {Grisoni}
\bibfield{author}{\bibinfo{person}{Francesca Grisoni},
  \bibinfo{person}{Claudia~S Neuhaus}, \bibinfo{person}{Miyabi Hishinuma},
  \bibinfo{person}{Gisela Gabernet}, \bibinfo{person}{Jan~A Hiss},
  \bibinfo{person}{Masaaki Kotera}, {and} \bibinfo{person}{Gisbert Schneider}.}
  \bibinfo{year}{[n.\,d.]}\natexlab{}.
\newblock \showarticletitle{De novo design of anticancer peptides by ensemble
  artificial neural networks}.
\newblock  (\bibinfo{year}{[n.\,d.]}).
\newblock
\urldef\tempurl%
\url{https://doi.org/10.1007/s00894-019-4007-6}
\showDOI{\tempurl}


\bibitem[Grover and Leskovec(2016)]%
        {grover2016node2vec}
\bibfield{author}{\bibinfo{person}{Aditya Grover} {and} \bibinfo{person}{Jure
  Leskovec}.} \bibinfo{year}{2016}\natexlab{}.
\newblock \showarticletitle{node2vec: Scalable feature learning for networks}.
  In \bibinfo{booktitle}{\emph{Proceedings of the 22nd ACM SIGKDD international
  conference on Knowledge discovery and data mining}}.
  \bibinfo{pages}{855--864}.
\newblock


\bibitem[Guo et~al\mbox{.}({[n.\,d.]})]%
        {Guo}
\bibfield{author}{\bibinfo{person}{Yuhang Guo}, \bibinfo{person}{Xiao Luo},
  \bibinfo{person}{Liang Chen}, {and} \bibinfo{person}{Minghua Deng}.}
  \bibinfo{year}{[n.\,d.]}\natexlab{}.
\newblock \showarticletitle{DNA-GCN: Graph convolutional networks for
  predicting DNA-protein binding}.
\newblock  (\bibinfo{year}{[n.\,d.]}).
\newblock
\urldef\tempurl%
\url{https://github.com/Tinard/dnagcn.}
\showURL{%
\tempurl}


\bibitem[Huang et~al\mbox{.}(2019)]%
        {huang2019explainable}
\bibfield{author}{\bibinfo{person}{Kexin Huang}, \bibinfo{person}{Cao Xiao},
  \bibinfo{person}{Lucas Glass}, {and} \bibinfo{person}{Jimeng Sun}.}
  \bibinfo{year}{2019}\natexlab{}.
\newblock \showarticletitle{Explainable substructure partition fingerprint for
  protein, drug, and more}. In \bibinfo{booktitle}{\emph{NeurIPS Learning
  Meaningful Representation of Life Workshop}}.
\newblock


\bibitem[Huang et~al\mbox{.}(2021)]%
        {Huang}
\bibfield{author}{\bibinfo{person}{Kai~Yao Huang}, \bibinfo{person}{Yi~Jhan
  Tseng}, \bibinfo{person}{Hui~Ju Kao}, \bibinfo{person}{Chia~Hung Chen},
  \bibinfo{person}{Hsiao~Hsiang Yang}, {and} \bibinfo{person}{Shun~Long Weng}.}
  \bibinfo{year}{2021}\natexlab{}.
\newblock \showarticletitle{Identification of subtypes of anticancer peptides
  based on sequential features and physicochemical properties}.
\newblock \bibinfo{journal}{\emph{Scientific Reports 2021 11:1}}
  \bibinfo{volume}{11} (\bibinfo{date}{6} \bibinfo{year}{2021}),
  \bibinfo{pages}{1--13}.
\newblock
Issue 1.
\showISBNx{0123456789}
\showISSN{2045-2322}
\urldef\tempurl%
\url{https://doi.org/10.1038/s41598-021-93124-9}
\showDOI{\tempurl}


\bibitem[Huang et~al\mbox{.}({[n.\,d.]})]%
        {Huangc}
\bibfield{author}{\bibinfo{person}{Lianzhe Huang}, \bibinfo{person}{Dehong Ma},
  \bibinfo{person}{Sujian Li}, \bibinfo{person}{Xiaodong Zhang}, {and}
  \bibinfo{person}{Houfeng Wang}.} \bibinfo{year}{[n.\,d.]}\natexlab{}.
\newblock \showarticletitle{Text Level Graph Neural Network for Text
  Classification}.
\newblock  (\bibinfo{year}{[n.\,d.]}).
\newblock
\urldef\tempurl%
\url{https://www.cs.umb.edu/}
\showURL{%
\tempurl}


\bibitem[Hwang et~al\mbox{.}(2019)]%
        {Hwang2019}
\bibfield{author}{\bibinfo{person}{Sohyun Hwang}, \bibinfo{person}{Chan~Yeong
  Kim}, \bibinfo{person}{Sunmo Yang}, \bibinfo{person}{Eiru Kim},
  \bibinfo{person}{Traver Hart}, \bibinfo{person}{Edward~M. Marcotte}, {and}
  \bibinfo{person}{Insuk Lee}.} \bibinfo{year}{2019}\natexlab{}.
\newblock \showarticletitle{HumanNet v2: human gene networks for disease
  research}.
\newblock \bibinfo{journal}{\emph{Nucleic acids research}}
  \bibinfo{volume}{47} (\bibinfo{date}{1} \bibinfo{year}{2019}),
  \bibinfo{pages}{D573--D580}.
\newblock
Issue D1.
\showISSN{1362-4962}
\urldef\tempurl%
\url{https://doi.org/10.1093/NAR/GKY1126}
\showDOI{\tempurl}


\bibitem[Islam et~al\mbox{.}(2021)]%
        {islam2021proximity}
\bibfield{author}{\bibinfo{person}{Muhammad~Ifte Islam},
  \bibinfo{person}{Farhan Tanvir}, \bibinfo{person}{Ginger Johnson},
  \bibinfo{person}{Esra Akbas}, {and} \bibinfo{person}{Mehmet~Emin Aktas}.}
  \bibinfo{year}{2021}\natexlab{}.
\newblock \showarticletitle{Proximity-based compression for network embedding}.
\newblock \bibinfo{journal}{\emph{Frontiers in big Data}}  \bibinfo{volume}{3}
  (\bibinfo{year}{2021}), \bibinfo{pages}{608043}.
\newblock


\bibitem[Kim et~al\mbox{.}({[n.\,d.]})]%
        {kim}
\bibfield{author}{\bibinfo{person}{Eun-Sol Kim}, \bibinfo{person}{† Woo},
  \bibinfo{person}{Young Kang}, \bibinfo{person}{Kyoung-Woon On},
  \bibinfo{person}{Yu-Jung Heo}, \bibinfo{person}{Byoung-Tak Zhang}, {and}
  \bibinfo{person}{Kakao Brain}.} \bibinfo{year}{[n.\,d.]}\natexlab{}.
\newblock \showarticletitle{Hypergraph Attention Networks for Multimodal
  Learning}.
\newblock  (\bibinfo{year}{[n.\,d.]}).
\newblock
\urldef\tempurl%
\url{https://spacy.io/}
\showURL{%
\tempurl}


\bibitem[Kuzmin et~al\mbox{.}(2020)]%
        {Kiril}
\bibfield{author}{\bibinfo{person}{Kiril Kuzmin},
  \bibinfo{person}{Ayotomiwa~Ezekiel Adeniyi}, \bibinfo{person}{Arthur~Kevin
  DaSouza}, \bibinfo{person}{Deuk Lim}, \bibinfo{person}{Huyen Nguyen},
  \bibinfo{person}{Nuria~Ramirez Molina}, \bibinfo{person}{Lanqiao Xiong},
  \bibinfo{person}{Irene~T. Weber}, {and} \bibinfo{person}{Robert~W.
  Harrison}.} \bibinfo{year}{2020}\natexlab{}.
\newblock \showarticletitle{Machine learning methods accurately predict host
  specificity of coronaviruses based on spike sequences alone}.
\newblock \bibinfo{journal}{\emph{Biochemical and Biophysical Research
  Communications}}  \bibinfo{volume}{533} (\bibinfo{date}{12}
  \bibinfo{year}{2020}), \bibinfo{pages}{553--558}.
\newblock
Issue 3.
\showISSN{0006-291X}
\urldef\tempurl%
\url{https://doi.org/10.1016/J.BBRC.2020.09.010}
\showDOI{\tempurl}


\bibitem[Leslie et~al\mbox{.}(2001)]%
        {leslie2001spectrum}
\bibfield{author}{\bibinfo{person}{Christina Leslie}, \bibinfo{person}{Eleazar
  Eskin}, {and} \bibinfo{person}{William~Stafford Noble}.}
  \bibinfo{year}{2001}\natexlab{}.
\newblock \showarticletitle{The spectrum kernel: A string kernel for SVM
  protein classification}.
\newblock In \bibinfo{booktitle}{\emph{Biocomputing 2002}}.
  \bibinfo{publisher}{World Scientific}, \bibinfo{pages}{564--575}.
\newblock


\bibitem[Narayanan et~al\mbox{.}(2017)]%
        {narayanan2017graph2vec}
\bibfield{author}{\bibinfo{person}{Annamalai Narayanan},
  \bibinfo{person}{Mahinthan Chandramohan}, \bibinfo{person}{Rajasekar
  Venkatesan}, \bibinfo{person}{Lihui Chen}, \bibinfo{person}{Yang Liu}, {and}
  \bibinfo{person}{Shantanu Jaiswal}.} \bibinfo{year}{2017}\natexlab{}.
\newblock \showarticletitle{graph2vec: Learning distributed representations of
  graphs}.
\newblock \bibinfo{journal}{\emph{arXiv preprint arXiv:1707.05005}}
  (\bibinfo{year}{2017}).
\newblock


\bibitem[Nguyen et~al\mbox{.}(2016)]%
        {Nguyen2016}
\bibfield{author}{\bibinfo{person}{Ngoc~Giang Nguyen}, \bibinfo{person}{Vu~Anh
  Tran}, \bibinfo{person}{Duc~Luu Ngo}, \bibinfo{person}{Dau Phan},
  \bibinfo{person}{Favorisen~Rosyking Lumbanraja},
  \bibinfo{person}{Mohammad~Reza Faisal}, \bibinfo{person}{Bahriddin Abapihi},
  \bibinfo{person}{Mamoru Kubo}, \bibinfo{person}{Kenji Satou},
  \bibinfo{person}{Ngoc~Giang Nguyen}, \bibinfo{person}{Vu~Anh Tran},
  \bibinfo{person}{Duc~Luu Ngo}, \bibinfo{person}{Dau Phan},
  \bibinfo{person}{Favorisen~Rosyking Lumbanraja},
  \bibinfo{person}{Mohammad~Reza Faisal}, \bibinfo{person}{Bahriddin Abapihi},
  \bibinfo{person}{Mamoru Kubo}, {and} \bibinfo{person}{Kenji Satou}.}
  \bibinfo{year}{2016}\natexlab{}.
\newblock \showarticletitle{DNA Sequence Classification by Convolutional Neural
  Network}.
\newblock \bibinfo{journal}{\emph{Journal of Biomedical Science and
  Engineering}}  \bibinfo{volume}{9} (\bibinfo{date}{4} \bibinfo{year}{2016}),
  \bibinfo{pages}{280--286}.
\newblock
Issue 5.
\showISSN{1937-6871}
\urldef\tempurl%
\url{https://doi.org/10.4236/JBISE.2016.95021}
\showDOI{\tempurl}


\bibitem[Pedregosa et~al\mbox{.}(2011)]%
        {scikit-learn}
\bibfield{author}{\bibinfo{person}{F. Pedregosa}, \bibinfo{person}{G.
  Varoquaux}, \bibinfo{person}{A. Gramfort}, \bibinfo{person}{V. Michel},
  \bibinfo{person}{B. Thirion}, \bibinfo{person}{O. Grisel},
  \bibinfo{person}{M. Blondel}, \bibinfo{person}{P. Prettenhofer},
  \bibinfo{person}{R. Weiss}, \bibinfo{person}{V. Dubourg}, \bibinfo{person}{J.
  Vanderplas}, \bibinfo{person}{A. Passos}, \bibinfo{person}{D. Cournapeau},
  \bibinfo{person}{M. Brucher}, \bibinfo{person}{M. Perrot}, {and}
  \bibinfo{person}{E. Duchesnay}.} \bibinfo{year}{2011}\natexlab{}.
\newblock \showarticletitle{Scikit-learn: Machine Learning in {P}ython}.
\newblock \bibinfo{journal}{\emph{Journal of Machine Learning Research}}
  \bibinfo{volume}{12} (\bibinfo{year}{2011}), \bibinfo{pages}{2825--2830}.
\newblock


\bibitem[Qian(2019)]%
        {Qian2019}
\bibfield{author}{\bibinfo{person}{Yujie. Qian}.}
  \bibinfo{year}{2019}\natexlab{}.
\newblock \showarticletitle{A graph-based framework for information
  extraction}.
\newblock  (\bibinfo{year}{2019}).
\newblock
\urldef\tempurl%
\url{https://dspace.mit.edu/handle/1721.1/122765}
\showURL{%
\tempurl}


\bibitem[Radicioni and Esposito(2010)]%
        {Radicioni}
\bibfield{author}{\bibinfo{person}{Daniele~P. Radicioni} {and}
  \bibinfo{person}{Roberto Esposito}.} \bibinfo{year}{2010}\natexlab{}.
\newblock \showarticletitle{BREVE: An HMPerceptron-based chord recognition
  system}.
\newblock \bibinfo{journal}{\emph{Studies in Computational Intelligence}}
  \bibinfo{volume}{274} (\bibinfo{year}{2010}), \bibinfo{pages}{143--164}.
\newblock
\showISBNx{9783642116735}
\showISSN{1860949X}
\urldef\tempurl%
\url{https://doi.org/10.1007/978-3-642-11674-2_7}
\showDOI{\tempurl}


\bibitem[Sahlin(2021)]%
        {sahlin2021strobemers}
\bibfield{author}{\bibinfo{person}{Kristoffer Sahlin}.}
  \bibinfo{year}{2021}\natexlab{}.
\newblock \showarticletitle{Strobemers: an alternative to k-mers for sequence
  comparison}.
\newblock \bibinfo{journal}{\emph{bioRxiv}} (\bibinfo{year}{2021}),
  \bibinfo{pages}{2021--01}.
\newblock


\bibitem[Saifuddin et~al\mbox{.}(2022)]%
        {saifuddin2022hygnn}
\bibfield{author}{\bibinfo{person}{Khaled~Mohammed Saifuddin},
  \bibinfo{person}{Bri Bumgardnerr}, \bibinfo{person}{Farhan Tanvir}, {and}
  \bibinfo{person}{Esra Akbas}.} \bibinfo{year}{2022}\natexlab{}.
\newblock \showarticletitle{HyGNN: Drug-Drug Interaction Prediction via
  Hypergraph Neural Network}.
\newblock \bibinfo{journal}{\emph{arXiv preprint arXiv:2206.12747}}
  (\bibinfo{year}{2022}).
\newblock


\bibitem[Saifuddin et~al\mbox{.}(2021)]%
        {Saifuddin2021}
\bibfield{author}{\bibinfo{person}{Khaled~Mohammed Saifuddin},
  \bibinfo{person}{Muhammad Ifte~Khairul Islam}, {and} \bibinfo{person}{Esra
  Akbas}.} \bibinfo{year}{2021}\natexlab{}.
\newblock \showarticletitle{Drug Abuse Detection in Twitter-sphere: Graph-Based
  Approach}.
\newblock \bibinfo{journal}{\emph{Proceedings - 2021 IEEE International
  Conference on Big Data, Big Data 2021}} (\bibinfo{year}{2021}),
  \bibinfo{pages}{4136--4145}.
\newblock
\showISBNx{9781665439022}
\urldef\tempurl%
\url{https://doi.org/10.1109/BIGDATA52589.2021.9671532}
\showDOI{\tempurl}


\bibitem[Saigo et~al\mbox{.}(2004)]%
        {Saigo2004}
\bibfield{author}{\bibinfo{person}{Hiroto Saigo},
  \bibinfo{person}{Jean~Philippe Vert}, \bibinfo{person}{Nobuhisa Ueda}, {and}
  \bibinfo{person}{Tatsuya Akutsu}.} \bibinfo{year}{2004}\natexlab{}.
\newblock \showarticletitle{Protein homology detection using string alignment
  kernels}.
\newblock \bibinfo{journal}{\emph{Bioinformatics (Oxford, England)}}
  \bibinfo{volume}{20} (\bibinfo{date}{7} \bibinfo{year}{2004}),
  \bibinfo{pages}{1682--1689}.
\newblock
Issue 11.
\showISSN{1367-4803}
\urldef\tempurl%
\url{https://doi.org/10.1093/BIOINFORMATICS/BTH141}
\showDOI{\tempurl}


\bibitem[Timmons and Hewage(2021)]%
        {Timmons2021}
\bibfield{author}{\bibinfo{person}{Patrick~Brendan Timmons} {and}
  \bibinfo{person}{Chandralal~M. Hewage}.} \bibinfo{year}{2021}\natexlab{}.
\newblock \showarticletitle{ENNAACT is a novel tool which employs neural
  networks for anticancer activity classification for therapeutic peptides}.
\newblock \bibinfo{journal}{\emph{Biomedicine \& pharmacotherapy = Biomedecine
  \& pharmacotherapie}}  \bibinfo{volume}{133} (\bibinfo{date}{1}
  \bibinfo{year}{2021}).
\newblock
\showISSN{1950-6007}
\urldef\tempurl%
\url{https://doi.org/10.1016/J.BIOPHA.2020.111051}
\showDOI{\tempurl}


\bibitem[Torrisi et~al\mbox{.}(2020)]%
        {Torrisi2020}
\bibfield{author}{\bibinfo{person}{Mirko Torrisi}, \bibinfo{person}{Gianluca
  Pollastri}, {and} \bibinfo{person}{Quan Le}.}
  \bibinfo{year}{2020}\natexlab{}.
\newblock \showarticletitle{Deep learning methods in protein structure
  prediction}.
\newblock \bibinfo{journal}{\emph{Computational and Structural Biotechnology
  Journal}}  \bibinfo{volume}{18} (\bibinfo{date}{1} \bibinfo{year}{2020}),
  \bibinfo{pages}{1301--1310}.
\newblock
\showISSN{2001-0370}
\urldef\tempurl%
\url{https://doi.org/10.1016/J.CSBJ.2019.12.011}
\showDOI{\tempurl}


\bibitem[Turner et~al\mbox{.}({[n.\,d.]})]%
        {Turner}
\bibfield{author}{\bibinfo{person}{Isaac Turner}, \bibinfo{person}{Kiran~V
  Garimella}, \bibinfo{person}{Zamin Iqbal}, {and} \bibinfo{person}{Gil
  Mcvean}.} \bibinfo{year}{[n.\,d.]}\natexlab{}.
\newblock \showarticletitle{Integrating long-range connectivity information
  into de Bruijn graphs}.
\newblock  (\bibinfo{year}{[n.\,d.]}).
\newblock
\urldef\tempurl%
\url{https://doi.org/10.1093/bioinformatics/bty157}
\showDOI{\tempurl}


\bibitem[UCI({[n.\,d.]})]%
        {UCI}
\bibfield{author}{\bibinfo{person}{UCI}.} \bibinfo{year}{[n.\,d.]}\natexlab{}.
\newblock \bibinfo{title}{UCI Machine Learning Repository: Data Sets}.
\newblock
\newblock
\urldef\tempurl%
\url{https://archive.ics.uci.edu/ml/datasets.php?format=&task=&att=&area=&numAtt=&numIns=&type=seq&sort=nameUp&view=table}
\showURL{%
\tempurl}


\bibitem[Vaswani et~al\mbox{.}(2017)]%
        {vaswani2017attention}
\bibfield{author}{\bibinfo{person}{Ashish Vaswani}, \bibinfo{person}{Noam
  Shazeer}, \bibinfo{person}{Niki Parmar}, \bibinfo{person}{Jakob Uszkoreit},
  \bibinfo{person}{Llion Jones}, \bibinfo{person}{Aidan~N Gomez},
  \bibinfo{person}{{\L}ukasz Kaiser}, {and} \bibinfo{person}{Illia
  Polosukhin}.} \bibinfo{year}{2017}\natexlab{}.
\newblock \showarticletitle{Attention is all you need}.
\newblock \bibinfo{journal}{\emph{Advances in neural information processing
  systems}}  \bibinfo{volume}{30} (\bibinfo{year}{2017}).
\newblock


\bibitem[Veli{\v{c}}kovi{\'c} et~al\mbox{.}(2017)]%
        {velivckovic2017graph}
\bibfield{author}{\bibinfo{person}{Petar Veli{\v{c}}kovi{\'c}},
  \bibinfo{person}{Guillem Cucurull}, \bibinfo{person}{Arantxa Casanova},
  \bibinfo{person}{Adriana Romero}, \bibinfo{person}{Pietro Lio}, {and}
  \bibinfo{person}{Yoshua Bengio}.} \bibinfo{year}{2017}\natexlab{}.
\newblock \showarticletitle{Graph attention networks}.
\newblock \bibinfo{journal}{\emph{arXiv preprint arXiv:1710.10903}}
  (\bibinfo{year}{2017}).
\newblock


\bibitem[Wang et~al\mbox{.}(2019)]%
        {Minjie}
\bibfield{author}{\bibinfo{person}{Minjie Wang}, \bibinfo{person}{Da Zheng},
  \bibinfo{person}{Zihao Ye}, \bibinfo{person}{Quan Gan},
  \bibinfo{person}{Mufei Li}, \bibinfo{person}{Xiang Song},
  \bibinfo{person}{Jinjing Zhou}, \bibinfo{person}{Chao Ma},
  \bibinfo{person}{Lingfan Yu}, \bibinfo{person}{Yu Gai},
  \bibinfo{person}{Tianjun Xiao}, \bibinfo{person}{Tong He},
  \bibinfo{person}{George Karypis}, \bibinfo{person}{Jinyang Li}, {and}
  \bibinfo{person}{Zheng Zhang}.} \bibinfo{year}{2019}\natexlab{}.
\newblock \showarticletitle{Deep Graph Library: A Graph-Centric,
  Highly-Performant Package for Graph Neural Networks}.
\newblock  (\bibinfo{date}{9} \bibinfo{year}{2019}).
\newblock
\urldef\tempurl%
\url{https://doi.org/10.48550/arxiv.1909.01315}
\showDOI{\tempurl}


\bibitem[Xie et~al\mbox{.}(2020)]%
        {Xie}
\bibfield{author}{\bibinfo{person}{Mingfeng Xie}, \bibinfo{person}{Dijia Liu},
  {and} \bibinfo{person}{Yufeng Yang}.} \bibinfo{year}{2020}\natexlab{}.
\newblock \showarticletitle{Anti-cancer peptides: classification, mechanism of
  action, reconstruction and modification}.
\newblock \bibinfo{journal}{\emph{Open Biology}}  \bibinfo{volume}{10}
  (\bibinfo{date}{7} \bibinfo{year}{2020}).
\newblock
Issue 7.
\showISSN{20462441}
\urldef\tempurl%
\url{https://doi.org/10.1098/RSOB.200004}
\showDOI{\tempurl}


\bibitem[Yao et~al\mbox{.}({[n.\,d.]})]%
        {Yaoa}
\bibfield{author}{\bibinfo{person}{Liang Yao}, \bibinfo{person}{Chengsheng
  Mao}, {and} \bibinfo{person}{Yuan Luo}.} \bibinfo{year}{[n.\,d.]}\natexlab{}.
\newblock \showarticletitle{Graph Convolutional Networks for Text
  Classification}.
\newblock  (\bibinfo{year}{[n.\,d.]}).
\newblock
\urldef\tempurl%
\url{www.aaai.org}
\showURL{%
\tempurl}


\bibitem[Zhang et~al\mbox{.}(2017)]%
        {zhang2017mining}
\bibfield{author}{\bibinfo{person}{Jingsong Zhang}, \bibinfo{person}{Jianmei
  Guo}, \bibinfo{person}{Xiaoqing Yu}, \bibinfo{person}{Xiangtian Yu},
  \bibinfo{person}{Weifeng Guo}, \bibinfo{person}{Tao Zeng}, {and}
  \bibinfo{person}{Luonan Chen}.} \bibinfo{year}{2017}\natexlab{}.
\newblock \showarticletitle{Mining k-mers of various lengths in biological
  sequences}. In \bibinfo{booktitle}{\emph{Bioinformatics Research and
  Applications: 13th International Symposium, ISBRA 2017, Honolulu, HI, USA,
  May 29--June 2, 2017, Proceedings 13}}. Springer, \bibinfo{pages}{186--195}.
\newblock


\bibitem[Zhang et~al\mbox{.}({[n.\,d.]})]%
        {Zhanga}
\bibfield{author}{\bibinfo{person}{Yufeng Zhang}, \bibinfo{person}{Xueli Yu},
  \bibinfo{person}{Zeyu Cui}, \bibinfo{person}{Shu Wu},
  \bibinfo{person}{Zhongzhen Wen}, {and} \bibinfo{person}{Liang Wang}.}
  \bibinfo{year}{[n.\,d.]}\natexlab{}.
\newblock \showarticletitle{Every Document Owns Its Structure: Inductive Text
  Classification via Graph Neural Networks}.
\newblock  (\bibinfo{year}{[n.\,d.]}).
\newblock
\showISBNx{2004.13826v2}
\urldef\tempurl%
\url{https://github.com/CRIPAC-DIG/TextING}
\showURL{%
\tempurl}


\bibitem[Zola(2014)]%
        {zola2014constructing}
\bibfield{author}{\bibinfo{person}{Jaroslaw Zola}.}
  \bibinfo{year}{2014}\natexlab{}.
\newblock \showarticletitle{Constructing similarity graphs from large-scale
  biological sequence collections}. In \bibinfo{booktitle}{\emph{2014 IEEE
  International Parallel \& Distributed Processing Symposium Workshops}}. IEEE,
  \bibinfo{pages}{500--507}.
\newblock


\end{thebibliography}
\twocolumn[\newpage]

\section{Appendix}
\subsection{ESPF and $k$-mer} \label{subsequence}
\textbf{ESPF:} ESPF stands for Explainable Substructure Partition Fingerprint. As for sub-word mining in the natural language processing domain, ESPF decomposes sequential inputs into a vocabulary list of interpretable moderate-sized subsequences. ESPF considers that a specific sequence property is mainly led by only a limited number of subsequences known as functional groups. We present the details of ESPF in Algorithm~\ref{algorithm:2}. 
Given a database, $S$ of sequences as input, ESPF generates a vocabulary list of subsequences as frequent reoccurring customized size subsequences. Starting with tokens as the initial set, it adds subsequences having a frequency above a threshold in $S$ to the vocabulary list. Subsequences appear in this vocabulary list in order of most frequent to least frequent. In our hypergraph, we use this vocabulary list as nodes and break down any sequence into a series of frequent subsequences relating to those. For any given sequence as input, we split it in order of frequency, starting from the highest frequency one.
An example of splitting a DNA sequence is as follows.

\begin{center}
\texttt{ GCTGAAAGCAACAGTGCAGACGATGAGACCGACGATCCCAGGAGGTAA }

$\Downarrow$

\texttt{ \underline{G} \underline{CTGAAAG} \underline{CAACAG} \underline{TGCAGA} \underline{CGA} \underline{TGAGA} \underline{CCGACGA} \underline{TCCCAG} \underline{GAGG} \underline{TAA}}
\end{center} 
\vspace{2mm}
\textbf{$\textbf{k}$-mer:} $k$-mer is an effective popular tool widely used in biological sequence data analysis (e.g., sequence matching). It splits sequential inputs into a list of overlapping subsequence strings of length $k$. To generate all $k$-mers from an input string, it starts with the first $k$ characters and then moves by just one character to get the next subsequence, and so on.  Steps for $k$-mer are shown in Algorithm~\ref{algorithm:3}.
If $t$ is the length of a sequence, there are $t-k+1$ numbers of $k$-mers and $T^k$ total possible number of $k$-mers, where $T$ is the number of monomers. $k$-mers can be considered as the words of a sentence. Like words, they help to attain the semantic features from a sequence. For example, for a sequence \texttt{ATGT},
monomers: \{A, T, and G\}, 2-mers: \{AT, TG, GT\}, 3-mers: \{ATG, TGT\}.

\SetKwComment{Comment}{/* }{ */}
\begin{algorithm}[h!]
\SetAlgoLined
\textbf{Input: } Initial Sequence tokens set ${S}$, tokenized Sequence strings set ${\tau}$, frequency threshold $\beta$, and size threshold ${M}$ for $S$.\\
 \For {${t = 1\dots,M}$}{
  $(a, b)$, $f$ $\leftarrow$ scan $\tau$ \Comment*[r]{$(a, b)$ is the frequentest consecutive tokens.}
  \If{$f < \beta$}
  {
   break \Comment*[r]{$(a, b)$'s frequency lower than threshold}
   }
   $\tau \leftarrow$ find $(a, b) \in $ $\tau$, replace with $(ab)$ \Comment*[r]{update $\tau$ with the combined token $(ab)$}
   $S \leftarrow S \cup (ab)$ \Comment*[r]{add $(ab)$ to the token vocabulary set $S$}
 }
\textbf{Output: } the updated tokenized drugs, $\tau$; the updated token vocabulary set, $S$.

\caption{Explainable Substructure Partition Fingerprint (ESPF)}
\label{algorithm:2} 
\end{algorithm}

\SetKwComment{Comment}{/* }{ */}
\begin{algorithm}[h!]
\SetAlgoLined
\textbf{Input: } $Sequences$, size threshold $k$\\
Subsequence\_list: []\\
Sequence\_dictionary:\{\}\\
\For {each $sequence$ in $Sequences$}{
LIST=[]\\
\For {$i$ in range ($t$-$k$+1)} 
{\tcc{$t$ is the length of $sequence$}
$D=sequence[i:i+k]$\\
$LIST.append(D)$\\
$Subsequence\_list.append(D)$
}
$Sequence\_dictionary[sequence]=LIST$
}
\textbf{Output: } Sequence\_dictionary, Subsequence\_list
\caption{$k$-mer}
\label{algorithm:3} 
\end{algorithm}

\subsection{Dataset Description}\label{dataset_description}
We evaluate the performance of our model using four different sequence datasets. 
They are (1) \textbf{Human DNA} sequence, (2) \textbf{Anti-cancer peptides}, (3) \textbf{Cov-S-Protein-Seq} and (4) \textbf{Bach choral} harmony. All these datasets are publicly available online. 
Table \ref{dataset_statistics} shows the statistics of the datasets.
\begin{table}[h!]
 \centering
    \caption{Statistics of Dataset}
      \normalsize
  \begin{tabular}{|c|c|c|c|}
    \hline
    \cellcolor{gray!60} \textbf{Dataset} &
     \cellcolor{gray!60} \textbf{\# of sample} &
     \cellcolor{gray!60} \textbf{avg. length} 
    \\
    \hline
    Human DNA & 4380 & 1264 \\
    \hline
    Bach choral & 5665 & 40  \\
    \hline
    Anticancer peptides & 939 & 17\\
    \hline
    Cov-S-Protein-Seq & 1238 &1297 \\
    \hline
  \end{tabular}
  \label{dataset_statistics}
\end{table}

1. The \textbf{Human DNA} (HD) sequence dataset consists of 4,380 DNA sequences. Each DNA sequence corresponds to a specific gene family (class), with a total of seven families: G protein-coupled receptors, Tyrosine Kinase, Tyrosine Phosphatase, Synthetase, Synthase, Ion channel, and Transcription factor. Our objective is to predict the gene family based on the coding sequence of the DNA. We obtain this dataset from Kaggle \cite{DNA}, which is a common online platform for machine learning and data scientists to collaborate between them.

2. The \textbf{Anti-cancer peptides} (ACPs) are short bioactive peptides typically formed of 10–60 amino acids \cite{Xie}. They have the characteristics to hinder tumor cell expansion and formation of tumor blood vessels and are less likely to cause drug resistance and low toxicity to normal cells \cite{Huang, Xie}. ACPs are found to interact with vital proteins to inhibit angiogenesis and recruit immune cells to kill cancer cells, such as HNP-110 \cite{Huang}. These unique advantages make ACPs the most promising anti-cancer candidate~\cite{Grisoni}. The ACP dataset contains 949 one-letter amino-acid sequences representing peptides and their four different anti-cancer activities (i.e., very active, moderately active, experimental inactive, virtual inactive) on breast and lung cancer cell lines \cite{UCI}. Given the amino-acid sequence, our goal is to predict the anti-cancer activities. 



3. The \textbf{Cov-S-Protein-Seq} (CPS) dataset consists of 1,238 spike protein sequences from 67 different coronavirus (CoV) species, including SARS-CoV-2 responsible for the COVID-19 pandemic \cite{Sarwan, Sandrine}. The dataset provides information on the CoV species (CVS) and their host species (CHS). The CoV species are grouped into seven categories, such as Avian coronavirus, Porcine epidemic diarrhea virus, Betacoronavirus1, Middle East respiratory syndrome coronavirus, Porcine coronavirus HKU 15, SARS CoV 2, and others. The host species are grouped into six categories: Swine, Avians, Humans, Bats, Camels, and other mammals. 
The goal is to predict the CoV species and host species based on the spike protein sequences. A cleaned and pre-processed version of the \textbf{Cov-S-Protein-Seq} dataset is used, collected from \cite{Kiril}.



4. Music is sequences of sounding events. Each event has a specific chord label. Chords are harmonic sets of pitches, also known as frequencies. The \textbf{Bach choral} harmony dataset is composed of 60 chorales having 5665 events \cite{Radicioni}. Each event of each chorale is labeled using one among 101 chord labels and described through 14 features, including 12 different pitch classes, Bass, and meter. Given this information, our target is to predict the chord label. In our experiment, out of 101 chord labels, we select the five most frequent chord labels. We get this dataset from UCI Machine Learning Repository \cite{UCI}.









\subsection{Number of Nodes and Edges in De-Bruijn Graphs} \label{node_edge_de_bruijn}
The number of nodes and edges in the De-Bruijn graphs for each dataset is presented in Table \ref{space_in_debruijn}. The selection of the $k$-value for constructing the De-Bruijn graph is based on the best performer of the \sht. For instance, in the case of the Human DNA dataset, a $k$-value of 25 yielded the best performance in the \sht; thus, a De-Bruijn graph is constructed using this value for this dataset.
\begin{table}[t!]
 \centering
    \caption{Number of Nodes (N) and Edges (E) in De-Bruijn Graphs}
      \normalsize
  \begin{tabular}{|c|c|c|c|}
    \hline
    \cellcolor{gray!60} \textbf{Dataset} &
     \cellcolor{gray!60} \textbf{\# of nodes} &
     \cellcolor{gray!60} \textbf{\# of edges} 
 \\
      & 
     $|N|$ &
     $|E|$ 
    \\
    \hline
    Human DNA & 5,422,447 &  5,418,151 \\
    \hline
    Bach choral & 34111 & 31890  \\
    \hline
    Anticancer peptides & 11939 & 11058\\
    \hline
    Cov-S-Protein-Seq & 1,576,019 &1,574,782 \\
    \hline
  \end{tabular}
  \label{space_in_debruijn}
\end{table}
\end{document}